\documentclass[pdflatex,sn-mathphys-num]{eta-sn-jnl}

\usepackage{graphicx}%
\usepackage{multirow}%
\usepackage{amsmath,amssymb,amsfonts}%
\usepackage{amsthm}%
\usepackage{mathrsfs}%
\usepackage[title]{appendix}%
\usepackage{xcolor}%
\usepackage{textcomp}%
\usepackage{manyfoot}%
\usepackage{booktabs}%
\usepackage{algorithm}%
\usepackage{algorithmicx}%
\usepackage{algpseudocode}%
\usepackage{listings}%
\usepackage{pdfpages}%

\theoremstyle{thmstyletwo}
\theoremstyle{thmstyleone}%

\theoremstyle{thmstyletwo}%

\theoremstyle{thmstylethree}%

\usepackage{dsfont}
\usepackage{gensymb}

\def\bfx{{\mathbf{x}}}

\def\bfu{{\mathbf{u}}}

\def\Re{\mathbb{R}}

\def\Pp{\mathbb{P}}

\def\Ee{\mathbb{E}}

\def\Bc{\mathcal{B}}

\def\Dc{\mathcal{D}}

\def\Fc{\mathcal{F}}

\def\Ic{\mathcal{I}}
\def\Jc{\mathcal{J}}

\def\Lc{\mathcal{L}}
\def\Mc{\mathcal{M}}
\def\Nc{\mathcal{N}}

\def\Pc{\mathcal{P}}
\def\Qc{\mathcal{Q}}

\def\Uc{\mathcal{U}}

\def\Xc{\mathcal{X}}
\def\Yc{\mathcal{Y}}

\def\ypush{y_{\#}}
\def\yfpush{y_{\xi\#}}
\def\yf{y_{\xi}}
\def\ygpush{y_{\eta\#}}
\def\yg{y_{\eta}}
\def\uf{u_\xi}
\def\ug{u_\eta}

\DeclareMathOperator*{\argmin}{arg\,min}

\usepackage{amsthm}
\theoremstyle{thmstyletwo}
\newtheorem{thm}{Theorem}
\theoremstyle{thmstylethree}%

\newtheorem{defn}[thm]{Definition}

\newtheorem{assump}[thm]{Assumption}

\usepackage{xcolor}

\definecolor{revisioncolor}{HTML}{B35C00}

\begin{document}

\title[Extreme Event Aware ($\eta$-) Learning]{Extreme Event Aware ($\eta$-) Learning}


\author[1,2]{\fnm{Kai} \sur{Chang}}\email{kaichang@mit.edu}

\author*[1]{\fnm{Themistoklis P.} \sur{Sapsis}}\email{sapsis@mit.edu}

\affil[1]{\orgdiv{Department of Mechanical Engineering}, \orgname{Massachusetts Institute of Technology}, \orgaddress{\street{77 Massachusetts Avenue}, \city{Cambridge}, \postcode{02139}, \state{MA}, \country{USA}}}

\affil[2]{\orgdiv{Center for Computational Science and Engineering}, \orgname{Massachusetts Institute of Technology}, \orgaddress{\street{77 Massachusetts Avenue}, \city{Cambridge}, \postcode{02139}, \state{MA}, \country{USA}}}


\abstract{Quantifying and predicting rare and extreme events is challenging because such events are infrequent, severe, and expensive to simulate. Existing data-driven methods often require multiple extremes in the training data or sampling process, leading to accurate predictions in quiescent regimes but high epistemic uncertainty in extreme-event regions. To overcome this limitation, we introduce Extreme Event Aware ($\eta$-) Learning, which does not require extreme events in the available data. The method reduces uncertainty even in uncharted extreme regimes by enforcing during training the statistics of an observable indicative of extremeness, obtained from qualitative knowledge or unlabeled data. This statistical regularization results in models that fit observed data while remaining consistent with prescribed observable statistics, enabling the generation of unprecedented extreme events. Optimal-transport-based theoretical results offer rigorous justification and establish key optimality properties. Numerical experiments on prototype systems and real-world precipitation downscaling problems demonstrate the effectiveness of the $\eta$-learning framework.}

\keywords{extreme events, statistical learning, generative modeling, optimal transport}



\maketitle

\section{Introduction}
The ubiquity of complex dynamical systems in science and engineering makes it crucial to accurately quantify and predict the behavior of these systems. To capture the intricacies in real-world scenarios, such systems are typically subject to some stochasticity. In these cases, fully understanding the statistical behavior of the systems requires knowledge of scenarios that rarely occur but induce significant consequences, i.e. extreme events \cite{sapsis2021statistics}. Common examples of extreme events include rogue ocean waves \cite{cousinsSapsis2015_JFM, dematteis2018rogue}, credit shocks \cite{fouque2011multiscale}, chemical transition states \cite{vanden2010transition}, and climate extremes \cite{seneviratne2021weather} to name just a few.

Quantifying and predicting extreme events is challenging due to two main obstacles. First, to accurately capture the extreme magnitudes in numerical simulations, it is necessary to have sufficiently high resolution as extreme events are characterized by local, strongly nonlinear and sharp transitions. However, high-fidelity simulations can take an impractical amount of time to run, and thus even a small number of them can be prohibitively expensive in practice \cite{sapsis2021statistics}. Second, since extreme events occur with a low probability, it is unlikely that they will appear until plentiful samples have been drawn. This makes it particularly challenging to accurately compute the statistics of the underlying dynamics, especially in the tail region. These two obstacles make standard simulation and sampling algorithms ineffective in characterizing extreme events \cite{sapsis2021statistics}.

For problems where one can simulate or perform experiments for any set of the system parameters, active learning methods have been formulated with the aim of judiciously choosing samples, to accelerate the convergence of extreme event statistics \cite{mohamad2018sequential, pickering2022discovering}. The main idea is to utilize the available data, build a probabilistic surrogate model, and decide where to sample next based on the surrogate and its uncertainty. By formulating optimal criteria that promote convergence of the tail, the extreme statistics can be accurately characterized with a very small number of samples \cite{sapsis2022optimal}. Despite the success of active learning, the implicit assumption is that the dataset contains at least a small number of data points associated with extreme events. These few data points, expressed through the surrogate, will steer the sampling process towards these extreme event regions, increase the accuracy there, and result in accurate tails. If the initial samples do not contain extreme events, then the active learning algorithm will randomly (thus inefficiently) explore the parameter space until it hits an extreme event region, i.e. it will not perform effectively. Ideas based on large deviations theory \cite{cousinsSapsis2015_JFM, dematteis2018rogue, tong2023large} are also limited by the same obstacles, i.e. they rely on the assumption that there exists a set of governing equations that are used to formulate an optimization problem that provides a set of informative realizations, the instantons. Solving the optimization problem is notoriously hard, similarly to active learning but perhaps more dramatically: since no surrogate is used, a high-dimensional space needs to be explored. The challenges associated with the lack of extreme event data are even more pronounced for problems where there is no capability to probe the system at specific parameter values, e.g. problems related to correction operators of coarse scale climate models, where training data are based exclusively on past observations \cite{barthel2024non, sorensen2024probabilistic}.

As it has become clear, all of the discussed methods, which are the state-of-the-art for the quantification of extremes, require knowledge of relevant extreme events or access through equations in some way to be useful. This necessity not only makes it prohibitive to employ the ML-based correctors when extreme data are not available in the first place, but it also stands as the bottleneck for the efficiency of sampling-based methods. Such a constraint leads us to ask the question: in the limited-extreme-data or no-extreme-data regime, is there a systematic way to see beyond the dataset and produce meaningful extremes?

In this work we formulate a framework to address this problem. We develop a theoretically justifiable learning paradigm that allows the generation of statistically consistent, physically plausible, and unprecedented extreme events without the assumption that the available data contain any extreme events. We call it extreme-event-aware (e2a or eta) or $\eta$-learning. At its core, $\eta$-learning is an optimization framework built upon two ingredients: 1) incorporating into the training process a priori available statistical information for an observable indicative of extremeness and 2) uncovering new, unencountered extreme events at inference time. Specifically, $\eta$-learning takes advantage of statistical information for a certain observable that is often available through, e.g., asymptotic analysis for weakly nonlinear systems \cite{Tayfun1980, Weinan1999,Soong_Grigoriou93}, white noise approximation for systems under stochastic excitation \cite{Sobczyk91, Belenky2019}, Gibbs hypothesis  \cite{Sullivan2009,lecun06,Zhu2019}, or simply through a dataset that consists of unlabeled data and cannot be used for training. However, even in cases where there is no such approximation available, one can hypothesize a distribution that can describe extreme events, e.g., a generalized extreme value distribution, and understand what type of realizations would be consistent with the underlying dataset and the assumed statistical distribution. In a nutshell, by properly incorporating prior statistical information into the workflow of data-driven models, $\eta$-learning allows for unveiling new scenarios that are physically meaningful and statistically consistent with the reference laws.

Beyond the physical intuitions, $\eta$-learning is also notable for its theoretical consistency. In particular, we develop a theory that provides an original perspective on how the push-forward measures of such estimators deviate from the ground truth in the limited-data regime. This is analyzed through the lens of optimal transport (OT) theory. The result demonstrates that without sufficient data or additional information, data-consistent estimators struggle to perform well in terms of output statistics. The formulation of $\eta$-learning, which extends the usual supervised learning, directly addresses this theoretical limitation, and the framework is theoretically justified. Moreover, by carefully analyzing the objective function of $\eta$-learning, we derive matching lower and upper bounds that not only offer clearer interpretations of its algorithmic behavior, but also substantiate the optimality in terms of minimizing the absolute distance between the $\eta$-map and the ground truth in expectation.


To make these ideas precise, we now introduce a mathematical formulation that captures the essential challenges of learning rare extremes in data-driven settings. At a high level, the setting considered here consists of three objects. First, an input \(\bfx\) describes the uncertain physical driver of a system, such as an initial condition, forcing parameter, or low-fidelity data such as low-resolution fields. Second, the system maps this input to a high-dimensional state \(u(\bfx)\), such as a dynamic trajectory or high-fidelity data. Third, an application-dependent observable \(g(u(\bfx))\) summarizes the state into a scalar quantity whose large values signify the extreme event of interest. Thus, the pipeline is
\[
\bfx \longmapsto u(\bfx) \longmapsto g(u(\bfx)).
\]
The learning task is to approximate either the full state map \(u\) or the composite response \(g\circ u\), even when the available training data contain few or no extreme events.

To formalize this setting, let \(X:(\Omega,\Sigma,P)\to(\Xc,\Bc(\Xc))\) be an input random variable, where \(\Xc\subset\mathbb{R}^d\) is equipped with its relative Borel \(\sigma\)-algebra. We denote the law of \(X\) by \(\mu\), namely
\[
\mu(A)=P(X^{-1}(A)), \qquad A\in\Bc(\Xc).
\]
We write \(X\sim\mu\), and use \(\bfx\sim\mu\) to denote a realization of the random input. Thus, \(\mu\) describes the distribution of inputs encountered by the system. We assume that \(\mu\) is either simple to sample from, such as a Gaussian distribution, or sufficiently well represented by existing samples.

Define the state map
\[
u:\Xc\to\Uc\subset\mathbb{R}^m,
\qquad
\bfx\mapsto u(\bfx),
\]
where \(\Uc\) is the state space. For each input realization \(\bfx\), \(u(\bfx)\) denotes the corresponding system state. In the settings of interest, accurate evaluations of \(u\) are computationally expensive, and available observations of \(u\) may be incomplete.

Next, define the pre-determined observable
\[
g:\Uc\to\Yc\subset\mathbb{R},
\qquad
\bfu\mapsto g(\bfu).
\]
This map provides a scalarization of the state space used in the subsequent distributional formulation. Examples include a spatial maximum, a spatial average, an energy functional, or another physically meaningful summary of the state. Throughout this work, we focus on scalar-valued observables; remarks on extensions to multivariate settings are provided in the Limitations and Future Work section.

Finally, let
\[
y:=g\circ u,
\qquad
Y:=y(X).
\]
Assuming \(y\) is \(\Bc(\Xc)\)-measurable with respect to \(\Bc(\Yc)\), the scalar response \(Y\) is a random variable with law given by the push-forward measure \(y_\#\mu\):
\[
y_\#\mu(B)=\mu(y^{-1}(B)), \qquad B\in\Bc(\Yc),
\]
where \(y^{-1}(B)=\{\bfx\in\Xc:y(\bfx)\in B\}\). We assume that \(y_\#\mu\) is absolutely continuous with respect to the Lebesgue measure on \(\mathbb{R}\), so that its probability density function exists.

As an example, consider the one-dimensional nonlinear dispersive wave turbulence equation introduced in \cite{majda1997one}. The input \(X\) may parameterize the initial condition through a Karhunen-Loève expansion \cite{pickering2022discovering}, the state \(u\) may be the real part of the corresponding solution, and the observable \(g\) may be the supremum norm \(g(\bfu)=\|\bfu\|_\infty\). In this case, evaluating \(y=g\circ u\) requires solving the governing equation with high fidelity, and large values of \(g\) naturally correspond to extreme wave events.

Mathematically, the mapping $y$ models how the output $Y$ reacts in response to the input $X$ through the system of interest. As the focus of this work lies in extreme events, we suppose that there exists an extreme set, $E \in \Bc(\Xc)$, and a threshold, $t_* \in \Yc$, such that
\begin{equation}
    \label{eq:extreme-set}
0 < \Pp\{X \in E\} = \Pp\{Y \geq t_*\} \leq \delta,
\end{equation}
where $\delta$, interpreted as a small quantity, denotes the maximal probability of $X$ falling into the extreme set $E$.

We study the data-driven setting in which we have access to a dataset \(\Dc=\{(\bfx_i,\bfu_i,y_i)\}_{i=1}^n\), where \(\bfu_i=u(\bfx_i)\) and \(y_i=g(\bfu_i)\). When the raw measurements used to construct \(\bfu_i\) may be noisy, \(\bfu_i\) should be understood as the processed training target obtained after any application-specific data-assimilation or bias-correction step. Such preprocessing requires additional assumptions, such as a measurement-noise model or an observation operator, and is therefore not a generic component of our setup. For the theoretical results below, the inputs \(\bfx_1,\ldots,\bfx_n\) are assumed to be independent and identically distributed according to \(\mu\). The extension to dependent data is provided in the Theoretical Extensions section of the Supplementary Information (SI).

Ideally, we would like to construct an estimator $\hat{u}$ from $\Dc$ so that the estimated push-forward density $(g\circ \hat{u})_\#\mu$ is close to the ground truth $y_\#\mu$, especially in the tail region wherein extreme events live. We use the phrases output statistics and push-forward measures interchangeably to refer to $\hat{y}_\#\mu$ and $y_\#\mu$. Depending on the specific application, either $\hat y$ or $\hat u$ will be the target. For instance, in Bayesian Optimization, often there is no need for explicit construction of $\hat{u}$ \cite{blanchard2021bayesian}, whereas for problems, e.g., in climate sciences, obtaining the full state function $u$ is indeed crucial, e.g., the full precipitation spatial field rather than its extreme value \cite{barthel2024non}.

Although the estimation problem above admits a natural supervised-learning formulation, our focus is on regimes where extreme events are intrinsic to the system but are rare or entirely absent from the available data. While there exists a vast literature on generalization error and statistical efficiency in supervised learning \cite{bach2024learning}, current frameworks do not provide targeted analyses of estimator performance in the presence of extreme events. This limitation is especially pronounced when the available dataset does not capture these events adequately. As a consequence, whether standard data-driven estimators can faithfully capture extreme-event statistics under data limitation remains understudied. In this work, we rigorously characterize this limitation and develop a principled remedy.

\section{Results}
Our first result characterizes a precise mathematical condition of data limitation under which a class of estimators fails to capture extreme events. Building on this, we subsequently describe the $\eta$-Learning framework and analyze its theoretical optimality in terms of capturing tail statistics. We then showcase the versatility and effectiveness of the \(\eta\)-learning algorithm by applying it to five distinct problems. Each of these problems is deliberately formulated under the assumption of data scarcity—a scenario where extreme events of interest are entirely absent from the dataset—to underscore the broad applicability of the \(\eta\)-learning framework.

\subsection*{Fundamental Limits of Learning Extreme Events under Data Scarcity}\label{sec:theory}
We begin by presenting a hardness theorem. We formalize a class of estimators, termed data-consistent estimators, that are more accurate in regions well-represented by the training data than in data-deficient regions, and argue that despite their wide adoption in practice, they are not ideal for capturing extreme statistics. Our result builds on OT theory and directly relates the distance between the true and approximate maps to the difference between the push-forward measures. The intermediate lemmas and proofs are postponed to the SI.

We note several assumptions in the current setting. First, both the dataset $\Dc$ and the estimator $\yf$ are treated as deterministic objects. This simplification is not an essential restriction of the argument. In the presence of training randomness of modern neural networks, the same argument applies conditioning on a realization of the randomness, while averaging over these sources gives the corresponding expected bound. Thus stochastic variance may change the realized constants or confidence level of the lower bound developed below, but it does not remove the missing-tail contribution caused by data scarcity. Second, all theoretical developments---including those in the next two sections---are derived for the full map \(y\), as opposed to the state map $u$. Third, we assume the absolute integrability of the relevant functions with respect to the corresponding probability measures, i.e. the existence of the first absolute moment. These limitations and considerations are either relaxed or addressed in the Theoretical Extensions section of the SI.

We first provide a guiding example, i.e. the Empirical Risk Minimization (ERM) estimator, of the types of estimators we consider in this work.
\begin{defn}[Empirical Risk Minimization (Chapter 4 of \cite{bach2024learning})]\label{defn:erm} Let $\Fc = \{f : \Xc \to \Yc\}$ be a family of predictors. A loss function $\ell$ that maps a predictor and a data point to a non-negative number is defined as
\begin{align*}
\ell : \Fc \times \Xc \times \Yc &\to \Re_{+} \cup \{0\}\\
(\phi, \bfx, y) &\mapsto \ell(\phi, \bfx, y).
\end{align*}
The ERM estimator $\yf$ is defined as
\begin{equation}
\yf := \argmin_{\phi\in \Fc }\frac{1}{n} \sum_{i=1}^n\ell(\phi, \bfx_i, y_i).
    \label{eq:erm}
\end{equation}
If the goal is to directly obtain an approximation to $u$, then $\Fc$ and $\ell$ are defined on $\Uc$ instead of $\Yc$. The ERM estimator reads
\begin{equation*}
    \yf := g \circ \uf,
\end{equation*}
where
\begin{equation}
\uf := \argmin_{\phi\in \Fc }\frac{1}{n} \sum_{i=1}^n\ell(\phi, \bfx_i, \bfu_i).
\label{eq:uerm}
\end{equation}
\end{defn}
A common example of the loss function $\ell$ defined above is the squared loss, that is,
\begin{equation}
\ell(\phi, \bfx, y) = (\phi(\bfx) - y)^2 \quad \text{or} \quad \ell(\phi, \bfx, \bfu) = \|\phi(\bfx) - \bfu\|_2^2.
    \label{eq:l2-loss}
\end{equation}
In this case, the ERM estimator is also known as the Mean-Squared-Error (MSE) estimator.

The theoretical argument aims at quantifying the difference between two probability measures. To do so, the first step is introducing a metric to measure the distance between them. We leverage OT theory and borrow the Wasserstein distance, which measures the least amount of work, in the sense of the Euclidean distance, required to transport one probability mass to the other \cite{sot}. In particular, we consider the 1-Wasserstein distance. The reasons for choosing the 1-Wasserstein distance will become clearer as we discuss its theoretical and practical advantages. It is defined as follows.

\begin{defn}[1-Wasserstein Distance \cite{sot}]\label{defn:w1}
    Let $\nu_1, \nu_2 \in \Pc_1(\Yc)$ be two probability measures, where $\Pc_1(\Yc)$ denotes the space of probability measures on $\Yc$ with a finite first moment. Let $\Pi(\nu_1, \nu_2)$ be the set of all probability couplings between $\nu_1$ and $\nu_2$; that is,
    \begin{align*}
            \Pi(\nu_1, \nu_2) := \{ &\gamma \in \Pc(\Yc \times \Yc) : \forall \text{ measurable } A \subset \Yc, \\
            &\gamma(A \times \Yc) = \nu_1(A),\, \gamma(\Yc \times A) = \nu_2(A) \}.
    \end{align*}
    The 1-Wasserstein distance between $\nu_1$ and $\nu_2$ is defined as
    \begin{equation}
    \label{eq:W1}
    W_1(\nu_1, \nu_2) = \inf_{\gamma \in \Pi(\nu_1, \nu_2)} \int \|y_1 - y_2\| \, \mathrm{d}\gamma(y_1, y_2),
    \end{equation}
    where $\|\cdot\|$ denotes the Euclidean norm.
\end{defn}

We now formalize the presence of extreme events from a function approximation perspective. In particular, our goal is to establish an interpretable mathematical condition that sheds light on why commonly used estimators struggle to capture extreme deviations when data are limited. The key insight underlying our approach is to directly connect the estimation error observed in the well-represented, non-extreme regions with that in the sparse, extreme regions. By bridging these error components, we expose a fundamental limitation: an estimator that performs well where data are abundant may still yield large discrepancies in the tails, thereby failing to approximate the true output statistics accurately.

\begin{defn}[Data-Consistent Estimator]\label{defn:data-consistent}
Let $\yf$ be an estimator selected from some function class to fit the dataset $\Dc$. Let $S$ be some smoothness parameter of the true mapping $y$. Recall that $E$ denotes the extreme set that satisfies the inequality \eqref{eq:extreme-set}. We say that $\yf$ is a data-consistent estimator if it satisfies either (or both) of the following conditions (i) and (ii). (i) When $\bfx_i \notin E$ for any $i$, we have
\begin{equation}
    \left|\int_{\Xc\setminus E}(y - \yf)\mathrm{d}\mu\right| \leq \tilde{C}(n, E, \Fc, S) \cdot \left|\int_E \left(y-\yf\right)\mathrm{d} \mu\right|,
    \label{eq:data-consistent}
\end{equation}
where
\begin{equation}
    \tilde{C} := \inf_{C\geq 0}\left\{\left|\int_{\Xc\setminus E}(y - \yf)\mathrm{d}\mu\right| \leq C \cdot \left|\int_E \left(y-\yf\right)\mathrm{d} \mu\right|\right\} < 1.
    \label{eq:ctilde}
\end{equation}
(ii) When $\bfx_i \notin E$ for any $i$, we have
\begin{equation}
    \int_{\Xc\setminus E}\left|y - \yf\right|\mathrm{d}\mu \leq \hat{C}(n, E, \Fc, S) \cdot \int_E \left|y-\yf\right|\mathrm{d} \mu,
    \label{eq:abs-data-consistent}
\end{equation}
where
\begin{equation}
    \hat{C} := \inf_{C\geq 0}\left\{\int_{\Xc\setminus E}\left|y - \yf\right|\mathrm{d}\mu \leq C \cdot \int_E \left|y-\yf\right|\mathrm{d} \mu\right\} < \infty.
    \label{eq:chat}
\end{equation}
Here, $\tilde{C}$ and $\hat{C}$ are some constants possibly depending on the function class $\Fc$, the smoothness of the true mapping (characterized by the parameter $S$), the number of data points $n$, and the extreme set $E$.
\end{defn}

The data-consistent conditions (i) and (ii) above, although related, do not imply one another. Their mathematical implications are clear: they both require an estimator \(\yf\) to exhibit larger expected errors in $E$, where data are absent, compared to $\Xc\setminus E$, where data are abundant. Although $\mu(E) < \mu(\Xc \setminus E)$ in general, this formulation captures two key features of the targeted problems: accurate approximation in the data-rich, non-extreme region and large errors in the data-scarce, extreme region despite its small measure.

We are now ready to state the following result: a lower bound on \(W_1\left(y_{\#} \mu, \yfpush \mu\right)\) for a {data-consistent} estimator.

\begin{thm}[Fundamental Limitation of Data Scarcity.]
    \label{thm:lower-bound}
    Suppose both $y$ and $\yf$ are in $L^1(\mu)$ and are $\Bc(\Xc)$-measurable with respect to $\Bc(\Yc)$. Then for a data-consistent estimator $\yf$ in the sense of \eqref{eq:data-consistent}, when $n\leq {\log p \over \log{(1-\delta)}}$, with probability at least $p$, we have
    \begin{equation}
        \label{eq:lower-bound}
        W_1\left(y_{\#} \mu, {\yfpush} \mu\right) \geq \left(1-\Tilde{C}(n, E, \Fc, S)\right) \left|\int_E \left(y-\yf\right)\mathrm{d} \mu\right|,
    \end{equation}
    where $\tilde{C}$ is the same as that in \eqref{eq:data-consistent}.
\end{thm}
Here, \( p \) represents a constant probability close to 1, while \( \delta \) is a small positive constant close to 0. The lower bound above provides valuable insight. It indicates that, for a data-consistent estimator, if the size of the training dataset falls below a certain threshold (i.e. in the data-deficient regime), then, with high probability, a gap always exists between the push-forward measures \( \yfpush \mu \) and \( \ypush \mu \). Specifically, the lower bound of this discrepancy, in terms of the \( W_1 \) metric, is primarily driven by the disparity between \( y \) and \( \yf \) in the extreme region \( E \), where training data is likely to be absent. When accurately capturing the tail is of particular interest, even a small deviation from \( y_\#\mu \) in $W_1$ can potentially result in a significant discrepancy in the tail. This limitation makes a data-consistent estimator inadequate for capturing the full statistical information of the true map in the context of extreme events, which leads us to the core methodological innovation in this work.

\subsection*{$\eta$-Learning Framework}\label{sec:framework}
Building on the limitations of data-consistent estimators established in Theorem \ref{thm:lower-bound}, we introduce the \(\eta\)-learning framework to tackle the challenge of discovering unprecedented extreme events when such occurrences are absent from the available data. While data-consistent estimators struggle to capture output statistics under data scarcity, \(\eta\)-learning is designed precisely to overcome this limitation while preserving the strengths of supervised learning in regions with sufficient data.

The core idea is to leverage a priori statistical information that is both physically meaningful and relevant to the problem at hand. By systematically incorporating this information into the supervised learning framework, the method enables statistically consistent learning of extreme events, ensuring that the resulting estimator becomes aware of and responsive to potentially critical extreme events, even when they are absent from the observed dataset. This approach integrates the predictive fidelity of supervised learning in well-sampled regions with the awareness of high-impact scenarios in data-scarce regions. As a particular example, we describe the framework based on the ERM loss, although the idea can be applied to any supervised learning framework.

Let \(\nu_0 \in \Pc_1(\Yc)\) be a probability measure on \(\Yc\) that encodes relevant statistical information about the extreme events of interest; its availability and construction are addressed in the Discussion section. This measure serves as a reference distribution that reflects the desired statistical characteristics of the output. $\eta$-learning integrates this information into the learning process by modifying the ERM objective to balance the fidelity of the observed data with alignment to \(\nu_0\).

When the primary goal is to approximate the mapping \(y\), $\eta$-learning formulates the following optimization problem:
\begin{equation}
\yg := \argmin_{\phi\in\Fc} \frac{1}{n} \sum_{i=1}^n\ell(\phi, \bfx_i, y_i) + \lambda \cdot W_1(\phi_\#\mu,\nu_0),
    \label{eq:geneces}
\end{equation}
where $\lambda >0$ is a balancing hyperparameter. Alternatively, if the aim is to model the underlying dynamics by approximating $u$ as in \eqref{eq:uerm}, $\eta$-learning modifies the optimization problem as follows:
\begin{equation}
\ug := \argmin_{\phi\in\Fc} \frac{1}{n} \sum_{i=1}^n\ell(\phi, \bfx_i, \bfu_i) + \lambda \cdot W_1((g\circ\phi)_\#\mu,\nu_0),
    \label{eq:geneces-u}
\end{equation}
and in this case, $\yg = g\circ \ug$.

To provide intuition for the effectiveness of \(\eta\)-learning, consider the ideal scenario where \(\nu_0 = y_\# \mu\). In this case, the ERM loss supervises \(\yg\) to capture the \(\bfx\)-\(y\) relationship in regions with abundant data, while the \(W_1\)-regularization term enforces alignment between \(\yg\)'s push-forward measure and the ground truth \(y_\# \mu\). This alignment directly addresses the challenges identified in Theorem \ref{thm:lower-bound}, where ERM struggles to capture the correct output statistics in data-scarce regimes. When the focus shifts to approximating \(u\) rather than \(y\), the $\eta$-learning framework retains its main advantage by ensuring that the learned dynamics not only approximate the observed data but also align with the statistical characteristics of \(\nu_0\) in the observable space. In this case, the optimization problem \eqref{eq:geneces-u} enforces the composite push-forward measure \((g \circ \ug)_\# \mu\) to match \(\nu_0\). If \(\nu_0 = y_\# \mu\), the framework ensures that the learned \(\ug\) captures the statistical features of \(y_\# \mu\), particularly the tails.

Beyond this intuitive explanation, we provide theoretical justification for the framework. Based on Theorem \ref{thm:lower-bound}, we establish that for a data-consistent estimator \(\yf\), the condition \(W_1(\yfpush \mu, y_\# \mu) \to 0\) is sufficient to ensure \(\yf \xrightarrow{\Pp} y\) on \(\Xc\), where \(\xrightarrow{\Pp}\) denotes convergence in probability. Although this result pertains to \(\yf\), it underscores the importance of output distribution alignment in achieving function-level convergence. This insight reinforces the rationale for incorporating the \(W_1\)-regularization term in \(\eta\)-learning: ultimately, our objective is to recover the mapping that exhibits extreme events.

We introduce the following additional assumption before formalizing the result.

\begin{assump}\label{assump:extreme-dom}
    $\int_E\left|y_{\xi}-y\right| d \mu \leq K \cdot\left|\int_E\left(y_{\xi}-y\right) d \mu\right|$ for some constant $K \ge 1$.
\end{assump}
This auxiliary assumption controls the possibility of cancellation in the error over the extreme set. It ensures that the integral error on $E$ does not vanish due to opposing positive and negative contributions, but rather reflects a systematic bias in the estimator. Intuitively, this corresponds to the regime where rare and extreme events are consistently misestimated (e.g., underestimated in magnitude), so that the signed error remains comparable to the absolute error.

\begin{thm}[Probabilistic Consistency.] \label{thm:sufficiency} In the setting of Theorem \ref{thm:lower-bound}, we have that if Definition \ref{defn:data-consistent} and Assumption \ref{assump:extreme-dom} are uniformly satisfied as $W_1(\yfpush \mu, y_\#\mu) \to 0$, then $\yf \xrightarrow{\Pp} y$ on $\Xc$.
\end{thm}

Theorem \ref{thm:sufficiency} reveals an important insight: effective recovery of extreme events is intrinsically linked to the distributional structure of the observable, justifying the \(W_1\) term in \(\eta\)-learning as a distributional regularizer in limited-data regimes. However, the condition that Assumption \ref{assump:extreme-dom} is uniformly satisfied as $W_1\left(y_{\xi \#} \mu, y_{\#} \mu\right) \rightarrow 0$ can be hard to impose during training without additional information. Therefore, if Assumption~5 cannot be assured, convergence in the observable distribution does not, by itself, imply point-wise reliability of the learned \(\eta\)-map. The learned map should then be interpreted as producing distributionally calibrated candidate extremes, rather than as guaranteeing recovery of their true location, time, or mechanism. Point-wise reliability requires additional identifying information, such as richer observables, spatial or temporal references, or structural and physical constraints during training.

Although Theorems \ref{thm:lower-bound} and \ref{thm:sufficiency} explicitly suggest incorporating the $W_1$ term in the loss function, it might seem that with the current arguments, alternative metrics on the space of probability distributions could also be considered. After all, minimizing any quantity that serves as an upper bound for \( W_1 \) would inherently force the \( W_1 \) distance to be reduced, which is still consistent with the theory following the same proof procedure of Theorems \ref{thm:lower-bound} and \ref{thm:sufficiency}. Examples of such metrics and divergences include the Kullback-Leibler (KL) divergence \cite{gibbs2002choosing} and the \( L^1 \)-log metric \cite{mohamad2018sequential}.

As it turns out, $W_1$ enjoys distinct advantages over alternative probability metrics, both theoretically and computationally. While computational considerations are deferred to the Discussion section, we show, through the next set of results, that $W_1$ is mathematically optimal for controlling tail errors under relevant assumptions. This property is particularly critical in our setting, where the primary motivation for introducing probabilistic regularization is to ensure the awareness of extreme events.

\subsection*{Optimality}\label{sec:optimality}
We now further analyze the optimization problem \eqref{eq:geneces} to show that both \(W_1(\yfpush \mu, y_\#\mu)\) and \(W_1(\ygpush \mu, y_\#\mu)\) are tight---hence optimal, up to multiplicative and additive constants---with respect to the expected approximation errors $\Ee\left[\left|(\yf-y)|_E\right|\right]$ and $\Ee\left[\left|(\yg-y)|_E\right|\right]$ in the extreme region. Specifically, under the idealized assumption $\nu_0 = y_\#\mu$, we derive matching upper and lower bounds that establish an equivalence between the $W_1$ regularization and tail error control. The intermediate lemmas and proofs are postponed to SI. We start by stating the result for \(W_1(\yfpush \mu, y_\#\mu)\).

\begin{thm}[Optimality of $W_1$ in Terms of Relative Approximation Errors.]\label{thm:optimality-rel}
    In the setting of Theorem \ref{thm:sufficiency}, we have
    \[
    \frac{1-\tilde{C}}{K}\int_E |\yf-y|\mathrm{d}\mu\leq W_1(\yfpush \mu, y_\#\mu)\leq (1+\hat{C})\int_E |\yf-y|\mathrm{d}\mu.
    \]
\end{thm}

Theorem \ref{thm:optimality-rel} provides a further justification of the $\eta$-learning framework. It indicates that, when the error induced by the data-consistent estimator is dominated by that occurs in the extreme region, the $W_1$ distance is, up to multiplicative constants, equivalent to the dominant error term $\Ee\left[\left|(\yf-y)\rvert_E\right|\right]$. This indicates that the incorporation of $W_1$ in \eqref{eq:geneces} and \eqref{eq:geneces-u} is optimal and that $W_1$ provides a particularly tight and informative representation of the tail error, rendering it optimally suited for the objectives of our framework.

Although Theorem \ref{thm:optimality-rel} further validates the framework by providing a certain optimality guarantee, similar to Theorem \ref{thm:sufficiency}, it does not directly apply to $\yg$. We next show that analogous tightness results on $\yg$ can be obtained in terms of exact approximation errors. To facilitate this, we introduce the following assumptions.

\renewcommand{\thethm}{8a}
\begin{assump}\label{assump:o1}
    $\int_{\Xc \setminus E} |\yf-y|\mathrm{d}\mu \leq \varepsilon_1.$
\end{assump}

\renewcommand{\thethm}{8b}
\begin{assump}\label{assump:o2}
    $\int_{\Xc \setminus E} |\yf-\yg|\mathrm{d}\mu \leq \varepsilon_2.$
\end{assump}
\setcounter{thm}{8}
\renewcommand{\thethm}{\arabic{thm}}

Intuitively, \(\varepsilon_1\) controls the residual error of the ordinary supervised estimator in the non-extreme region, where labeled data are abundant. It therefore represents the baseline bulk approximation error. The parameter \(\varepsilon_2\) controls the deviation of the \(\eta\)-estimator from this supervised estimator in the same non-extreme region. Hence, Assumption~\ref{assump:o1} requires the baseline estimator to be reasonably accurate in the non-extreme region, while Assumption~\ref{assump:o2} further requires the discrepancy between the estimators \(\yf\) and \(\yg\) to be controlled there. These are natural conditions because both estimators are constrained by the same supervised-learning objective on the data-rich non-extreme region. With these assumptions in place, we are prepared to derive bounds expressed in terms of the exact approximation errors.

\begin{thm}[Optimality of $W_1$ in Terms of Exact Approximation Error.]\label{thm:optimality-exact}
    Suppose $\int_E\left|y_{\eta}-y\right| d \mu \leq K \cdot\left|\int_E\left(y_{\eta}-y\right) d \mu\right|$ for some constant $K \ge 1$. In the setting of Theorem \ref{thm:optimality-rel}, under Assumptions \ref{assump:o1} and \ref{assump:o2}, as $K \to 1$, we have
    \[
    \left| W_1(\ygpush \mu, y_\#\mu) - \int_E |\yg-y|\mathrm{d}\mu \right| \leq \varepsilon_1 + \varepsilon_2.
    \]
\end{thm}
Theorem \ref{thm:optimality-exact} establishes that, up to approximation errors induced by the data-consistent estimator in the non-extreme region—where data are abundant—and up to the discrepancy between the data-consistent estimator and the \(\eta\)-estimator in this region, \(W_1(\ygpush \mu, y_\#\mu)\) is effectively equivalent to \(\int_E |\yg - y| \, \mathrm{d}\mu\), which represents the dominant approximation error associated with extreme events. To further interpret this result, recall that \(\yg\) is defined to minimize \(W_1(\ygpush \mu, y_\#\mu)\). Given this equivalence, minimizing \(W_1(\ygpush \mu, y_\#\mu)\) inherently leads to the minimization of \(\int_E |\yg - y| \, \mathrm{d}\mu\). This result directly confirms the effectiveness of \(\eta\)-learning in capturing extreme behaviors while offering a clearer interpretation of its algorithmic behavior: in essence, the \(\eta\)-estimator implicitly seeks to align itself with the true mapping \(y\) in expectation, particularly within regions characterized by extreme events, when the associated assumptions are satisfied.

\subsection*{A 2D-to-1D Toy Problem}
The numerical results presented henceforth serve as proof-of-concept validations of the \(\eta\)-learning method. Specifically, we consistently operate under the ideal assumption that \(\nu_0 = y_\# \mu\) attaching to our developed theory if not explicitly mentioned otherwise. The process of obtaining a certified \(\nu_0\), if not available beforehand, is application-specific and thus remains an area for future exploration. Additionally, we employ the widely adopted quadratic loss function defined in equation \eqref{eq:l2-loss}. The ERM estimator will thus be referred to as the Mean-Squared-Error (MSE) estimator. The function class $\Fc_\theta$ in both \eqref{eq:geneces} and \eqref{eq:geneces-u} is taken as a neural network (NN) \cite{goodfellow2016deep} for all cases. We also comment that we refer to PDFs as being approximated from well-calibrated, high-fidelity kernel density estimators (KDEs) using standard packages (e.g. \texttt{scipy.stats} \cite{2020SciPy-NMeth}). All other relevant details are included in the SI under the respective sections.

\begin{figure*}[t]
    \centering
    \includegraphics[width=.95\linewidth]{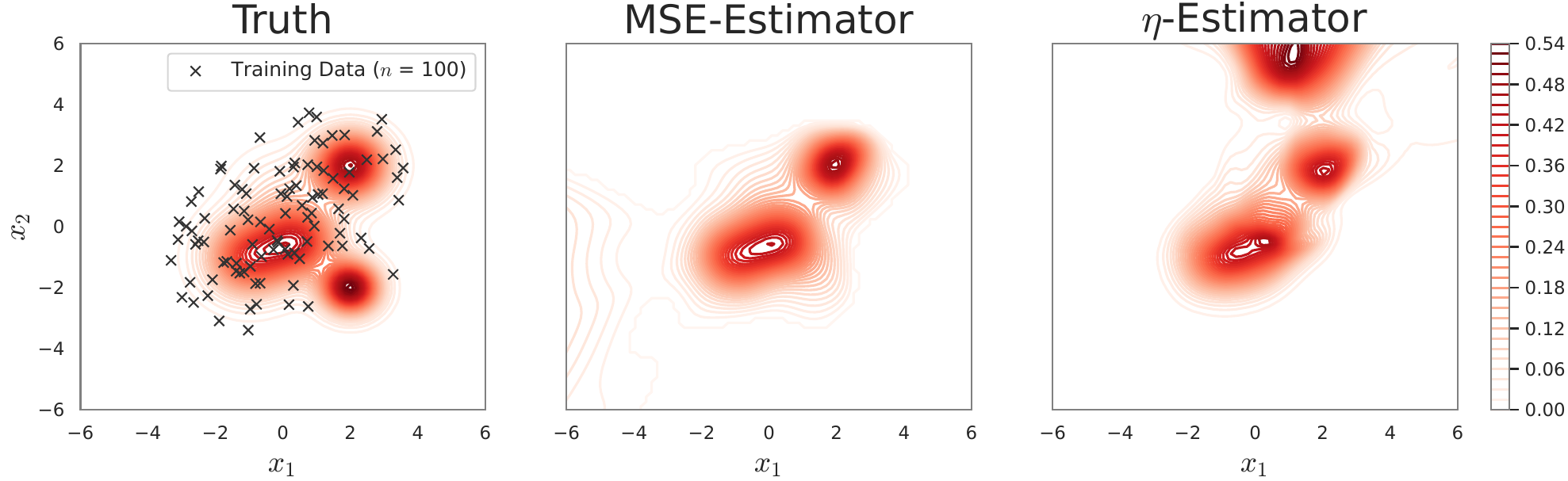}
    \caption{Comparison of estimators for the 2D-to-1D toy problem. Contours are evaluated on $[-6,6] \times [-6,6]$. Left: ground-truth map, with crosses marking training data that do not contain the extreme of interest. Middle: MSE estimator trained with the selected training data. Right: $\eta$-estimator trained with the same data and a reference distribution.}
    \label{fig:toy1d-map-compare}
\end{figure*}

We begin with a toy problem for \eqref{eq:geneces}, aimed at directly approximating the nonlinear map $y$. Let $\bfx \stackrel{d}{=} X \in \Re^2$ be an isotropic Gaussian random variable, $X \sim \mu = \Nc(0,\sigma^2I)$, where \( \sigma^2 = 10 \). The true mapping is constructed as a superposition of multiple Gaussian functions, one of which, with the smallest variance, is situated in the tail region. This particular peak represents the extreme of interest. The visualization of $y$ is shown in Fig. \ref{fig:toy1d-map-compare}, with its explicit formula stated in the SI. The training data are constructed by sampling $n=100$ points in the input space, explicitly bypassing the targeted extreme region, as illustrated in the leftmost plot of Fig. \ref{fig:toy1d-map-compare}. This setup reflects a data-limited regime.

Fig.~\ref{fig:toy1d-map-compare} shows the true map, the MSE map, and one realization of the \(\eta\)-map, while Supplementary Fig.~1 compares the induced distributions \(y_\#\mu\), \(\ygpush \mu\), and \(\yfpush \mu\). Additional \(\eta\)-map realizations and their corresponding PDF comparisons are shown in SI Figs. 2 and 3.  The MSE map accurately fits the data-rich region but fails to infer the unobserved target extreme region, illustrating the data-consistent setting of Definition~\ref{defn:data-consistent} and the fundamental limitation of MSE-based learning under tail data scarcity. In contrast, the \(\eta\)-map does not exactly recover the true extreme structure, but it introduces high-output mass in the tail while preserving the prediction in the data-rich region. This behavior is expected by theory: because the assumptions of Theorem~\ref{thm:sufficiency} are not enforced as constraints during training, exact pointwise recovery is not guaranteed, and the available guarantee is convergence in \(L^1(\mu)\). Distributionally, however, Supplementary Fig.~1 shows that \(\ygpush \mu\) is substantially closer to the reference distribution, which is the truth in this example, than the MSE-induced law, especially in the tail. Thus, \(\eta\)-learning reduces the epistemic uncertainty caused by missing extreme samples by producing statistically plausible extremes, rather than by uniquely identifying their exact spatial configuration.

The additional realizations in Supplementary Fig.~2 show that distributional convergence does not uniquely determine the spatial location of the inferred high-output region. We quantify this variability in the SI by training \(K=20\) independently initialized \(\eta\)-estimators under the same configuration. The resulting maximizer locations have ensemble spatial standard deviation \(\sigma_{\mathrm{loc}}=3.992\), while the 2-norm of the true maximizer is 2.86. Thus, independently initialized \(\eta\)-estimators can place the inferred high-output region at substantially different spatial locations.

The ensemble analysis highlights a broader implication of $\eta$-learning: enforcing scalar observable tail statistics generates candidate extremes, but does not necessarily identify the underlying physical structure, such as the peak location shown in this example. This variability reflects the residual physical uncertainty that remains after enforcing the observable tail statistics, rather than a failure to recover a uniquely identifiable hidden event. Nevertheless, two aspects are stable across realizations: predictions remain accurate in the data-rich region, and, as shown in Supplementary Fig.~3, the induced output PDFs remain consistent with the reference distribution. In the Discussion section, we further describe how this uncertainty can be systematically characterized and used in downstream applications.

Additionally, as shown in Supplementary Fig.~4, $\ygpush \mu$ remains stable as the balancing hyperparameter $\lambda$ varies from $10^{-4}$ to $10$, spanning five orders of magnitude, while all other training configurations are fixed. This indicates that our method is numerically robust to the choice of $\lambda$, a practically desirable property. This robustness is enabled by the staged training procedure described later in the Effective Training Strategies subsection.

\subsection*{A 2D-to-2D Toy Problem}
\begin{figure}[t]
    \centering
    \includegraphics[width=.95\linewidth]{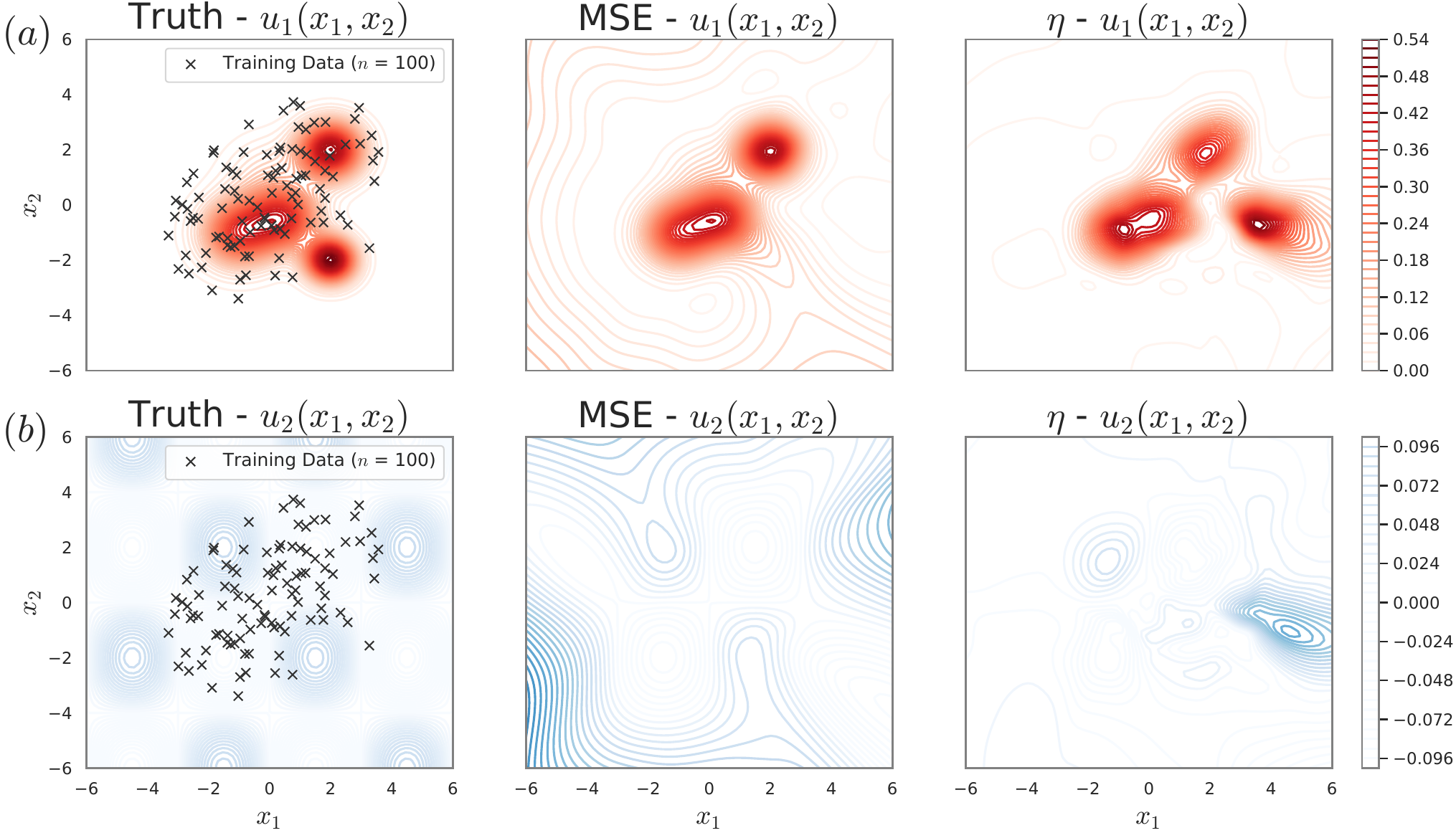}
    \caption{Comparison of estimators for the 2D-to-2D toy problem. Contours are evaluated on $[-6,6] \times [-6,6]$. (a) First output component, $u_1$. (b) Second output component, $u_2$. In each panel, the three maps correspond to the ground-truth map, with crosses marking training data that do not contain the extreme of interest; the MSE estimator trained with those data; and the $\eta$-estimator trained with the same data and a reference distribution, respectively.}
    \label{fig:toy2d-map-compare}
\end{figure}

Next, we consider a toy example with a highly nonlinear intermediate state function $u : \Re^2 \to \Re^2$. The input distribution is the same as in the previous case. The specifics of $u$ and the observable $g$ are detailed in the SI. A visualization of $u$ and the training data is provided in Fig. \ref{fig:toy2d-map-compare}. The training setup remains consistent with the 2D-to-1D example, except that the output dimension of the NN is 2D.

Fig.~\ref{fig:toy2d-map-compare} shows the true map, the MSE map, and one realization of the \(\eta\)-map, while Supplementary Fig.~5 compares the induced observable distributions \(y_\#\mu\), \(\ygpush\mu\), and \(\yfpush\mu\). Additional \(\eta\)-map realizations and their corresponding PDF comparisons are shown in SI Figs.~6 and 7. As in the previous example, the MSE estimator accurately captures the map only in the data-rich region, but fails to detect the unobserved target extreme region. In contrast, the \(\eta\)-map uses the reference observable distribution to generate high-output responses in the missing tail region. Although the reconstructed extremes are not pointwise exact, the induced observable PDF remains closely aligned with the reference distribution, as shown in Supplementary Fig.~5.

This example also exposes a limitation of constraining a scalar observable for high-dimensional states. Since $g(\bfu)=2|u_1|+\frac{1}{2}|u_2|$ aggregates the two state components into a single scalar quantity, the \(W_1\) penalty constrains the observable law, but not how the extreme response decomposes between the two components. Compared with the 2D-to-1D case, this introduces an additional ambiguity: distortion in one state variable can be partially compensated by changes in the other while preserving similar scalar observable statistics. This scalarization-induced interference explains the more pronounced spatial distortion visible in Fig.~\ref{fig:toy2d-map-compare}, especially in \(u_1\).

The additional realizations in Supplementary Fig.~6 further show variability in the inferred extreme-region structure across neural-network initializations. Nevertheless, the induced observable PDFs remain consistently aligned with the reference distribution, as shown in Supplementary Fig.~7. Overall, these results demonstrate that \(\eta\)-learning can generate plausible tail events under missing extreme data, but scalar-observable matching alone may leave residual uncertainty in the component-wise structure of high-dimensional states.

\subsection*{A Real-World Precipitation Downscaling Challenge}

Starting from this example, we examine three applications of $\eta$-learning related to a challenging real-world precipitation downscaling problem. In statistical downscaling, the challenge of data scarcity arises from the disparity between low-resolution (LR) simulation data, which typically span significantly longer time periods, and high-resolution (HR) observational data, which are particularly limited in temporal coverage. The simulation data is often generated at a spatial resolution insufficient to capture the finer details (extremes) of climate dynamics. As a result, the statistics of the observable for the coarser simulations typically deviate substantially due to unresolved dynamics \cite{barthel2024non}. Statistical downscaling seeks to address this issue by enhancing the spatial resolution of under-resolved data while ensuring that the observable statistics, particularly in the extreme tail regions, align with the ground truth.

We begin with the standard statistical downscaling task on the chosen precipitation data. We use the ERA5-Land dataset \cite{munoz2021era5} and download 25 years of hourly total precipitation (the variable \texttt{tp}) data from 1999 to 2023 on the main continent of the United States. Details of data processing are listed in the SI. To understand how statistical downscaling fits into the $\eta$-learning framework, consider $\bfx$, the input, as the LR fields and $\bfu$, the state information, as the HR fields. The observable $g$ is selected as the spatial maximum of a precipitation field, $\max(\cdot)$, a natural candidate for studying extremes \cite{coles2001introduction}, with the maximum taken over the full domain for each realization (i.e. snapshot) of $\bfu$. We thus also write $\Dc_{\text{lo}} = \{\bfx_i\}_{i=1}^{9044} \sim \mu$ and $\Dc_{\text{hi}} = \{\bfu_i\}_{i=1}^{9044}$. Consequently, the downscaling function can be naturally modeled as an $\ug$-map.

\begin{figure*}[t]
    \centering
    \includegraphics[width=.95\linewidth]{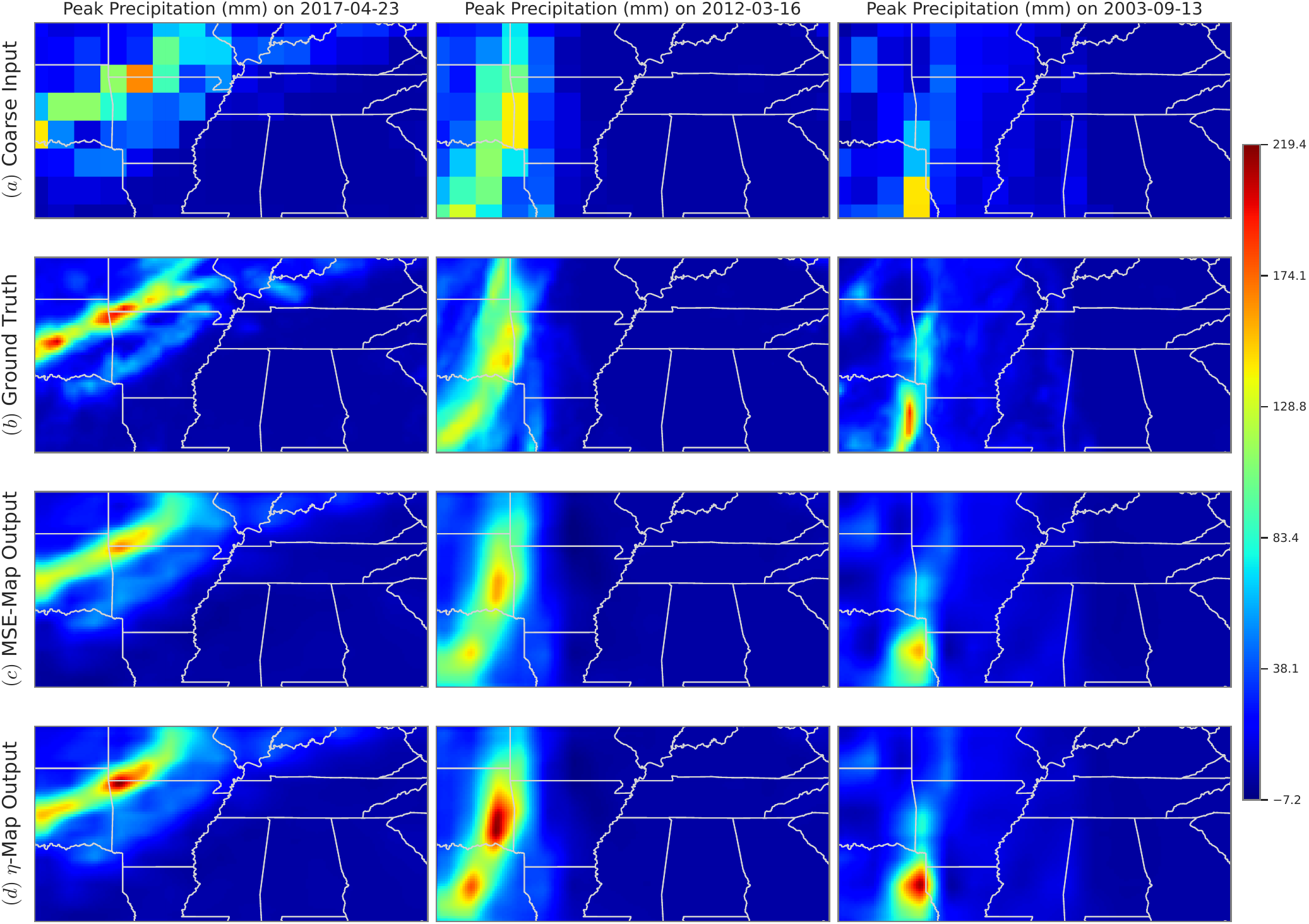}
    \caption{Representative precipitation downscaling snapshots. Each column corresponds to a day in the test set. (a) Low-resolution (LR) precipitation samples. (b) Ground-truth high-resolution (HR) samples. (c) HR samples obtained by super-resolving the LR samples in (a) through the MSE map. (d) HR samples obtained by super-resolving the LR samples in (a) through the $\eta$-map.}
    \label{fig:sample-downscale}
\end{figure*}

\begin{figure}
    \centering
    \includegraphics[width=.99\linewidth]{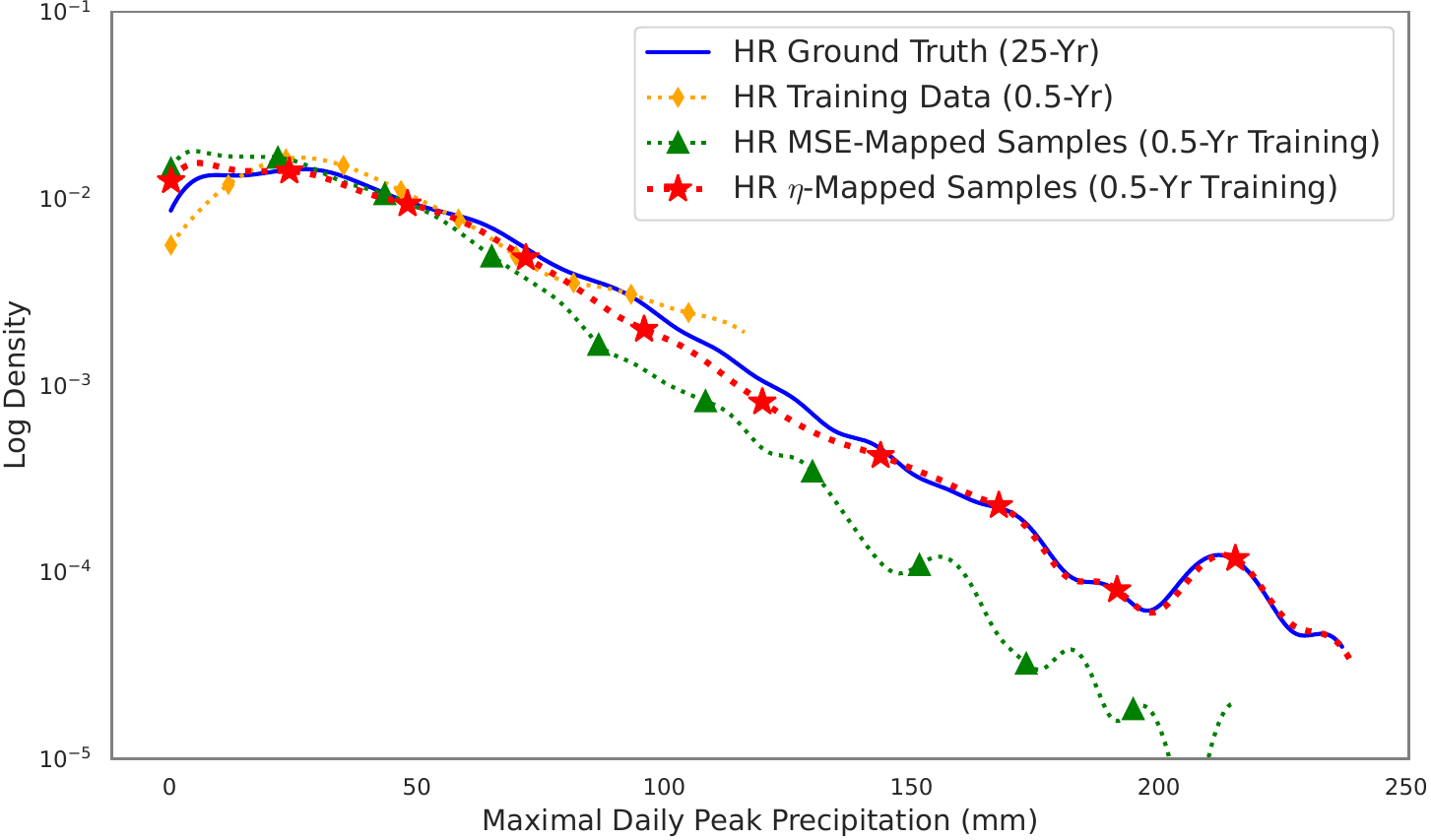}
    \caption{Probability densities of high-resolution (HR) daily maximal peak precipitation under different processing methods. Solid blue curve: 25-year ground-truth HR data. Dotted orange curve with rhombus markers: 0.5-year HR training data. Dotted green curve with triangle markers: HR samples obtained by super-resolving 25-year low-resolution (LR) samples through the MSE map. Dotted red curve with star markers: HR samples obtained by super-resolving 25-year LR samples through the $\eta$-map.}
    \label{fig:pdf-downscale}
\end{figure}

The visualization of several selected tail samples of the precipitation fields is presented in Fig. \ref{fig:sample-downscale}, with each column representing a different sample and each row corresponding to a particular processing method (more details are provided in the caption). In the left and right examples, we observe that while the $\uf$-downscaled HR fields (third row) largely miss the extreme events, the $\ug$-downscaled HR fields (last row) generate statistically consistent high-magnitude fields, albeit with imperfect spatial localization. This discrepancy arises because $\ug$ is informed solely by the observable PDF information and does not consider explicit spatial guidance in the unobserved extreme regime. In the middle example, the $\ug$-downscaled HR field exhibits an extreme event while the $\uf$-downscaled field does not; however, the ground truth (second row) does not manifest any extreme event. Rather than indicating a methodological failure, this outcome underscores the probabilistic nature of the $\eta$-map: the distribution-matching regularization does not specify the exact spatial occurrence of extremes in the absence of supplementary spatial information. It is also crucial to note that the dynamical structure in the generated extreme events does not come from the Wasserstein regularization alone. Rather, it is inherited from the supervised problem: each LR snapshot provides a coarse representation of the corresponding physical state, while the paired LR-HR snapshots over the first 0.5 years provide labeled spatial correspondence. The key point is that the distributional information induced by $\nu_0$ is well captured by $\ug$, as can be seen in Fig. \ref{fig:pdf-downscale}. In that figure, it is evident that while $\yfpush \mu$ completely misses the true distribution, $\ygpush\mu$ closely matches the tail by learning only the PDF information beyond the quantile value of 150. Interestingly, the tail correction provided by $\ygpush\mu$ also improves the representation of the bulk distribution compared to $\yfpush \mu$. Overall, these results demonstrate that the $\eta$-map effectively generates unencountered, physically plausible, and extreme scenarios that are consistent with the prescribed PDF information.

We further evaluate two statistics that are not directly enforced during training: the conditional mean $m(\bfu_0; t, n_g)$ and the weighted coverage $c(\bfu_0; t, n_g)$, where $\bfu_0$ is a sample field, \(t\) is a threshold, and \(n_g\) is the number of spatial grid cells. The conditional mean measures the average magnitude above a threshold, while the weighted coverage measures the fraction of total field magnitude contributed by entries exceeding that threshold. Formal definitions are given in the SI. For $m$, we compute the PDFs across a range of thresholds from 154 to 214, and similarly for $c$. As shown in Supplementary Fig. 8, the $\ug$-mapped samples consistently outperform the $\uf$-mapped samples with respect to the conditional mean statistic, further confirming the effectiveness of $\eta$-learning in capturing extreme values. Supplementary Fig.~9 shows a trade-off at lower thresholds: the \(\ug\)-mapped weighted-coverage distribution agrees less well with the truth than the $\uf$-mapped distribution, especially in the bulk. This is expected because the training objective constrains the spatial maximum, not the magnitude-weighted exceedance fraction. However, as the threshold increases, the MSE-PDF progressively collapses into a Dirac delta, reflecting its failure to represent extremes, whereas the $\eta$-PDF continues to provide a valid proxy of the true distribution.

To further quantify the associated full-field cost, we provide RMSE and structural similarity index measure (SSIM) diagnostics in SI Table~1. Over all \(9044\) paired fields, the full-grid RMSE increases from \(2.878\) for $\uf$ to \(3.112\) for $\ug$, an \(8.13\%\) relative increase. Average SSIM ($\bar{\mathrm{SSIM}}$) changes only from \(0.947\) to \(0.944\), a \(0.30\%\) relative decrease. We also evaluate subsets stratified by $\nu_0$: define \(\Ic_{\le q}=\{1\le j\le 9044:\max (\bfu_j)\le F_{\nu_0}^{-1}(q)\}\) and $\Ic_{\ge q}$ similarly, where $F_{\nu_0}^{-1}(q)$ denotes the $q$-quantile of $\nu_0$. For bulk subsets $\Ic_{\le 0.7}$, $\Ic_{\le 0.8}$, and $\Ic_{\le 0.9}$, the \(\ug\)-RMSE increase ranges from \(6.20\%\) to \(6.92\%\), while the $\ug$-$\overline{\mathrm{SSIM}}$ decrease remains below \(0.27\%\). For tail subsets $\Ic_{\ge 0.95}$, $\Ic_{\ge 0.975}$, and $\Ic_{\ge 0.99}$, the $\ug$-RMSE increase ranges from \(9.21\%\) to \(10.30\%\), while the $\ug$-$\overline{\mathrm{SSIM}}$ decrease remains below \(0.48\%\). These results show that emphasizing the scalar observable introduces a modest pointwise-intensity cost, with a larger effect on high-maximum fields, but does not produce a significant degradation in the local spatial structure measured by SSIM. Additional details of the spatial-fidelity analysis are provided in the SI.

Taken together, these diagnostics clarify the trade-off. On one hand, \(\eta\)-learning captures the prescribed extreme-observable statistics and transfers this improvement to related high-threshold statistics that are not explicitly enforced. On the other hand, this improvement comes with a modest spatial fidelity cost, primarily through pointwise amplitudes rather than a broad degradation of spatial morphology.

\subsection*{Correcting Statistical Biases of Deep Generative Models}

In the next example, we demonstrate how to utilize \( \eta \)-learning as a model-agnostic statistical corrector for deep generative models (DGMs), making it applicable to any DGM framework. We choose Flow Matching (FM) \cite{lipman2022flow} as our DGM. The aim is to generate, in principle, an arbitrary number of HR samples from the underlying data manifold characterized by the observable distribution. Since DGMs typically require large amounts of data for effective training, this approach is particularly valuable when the available HR data are insufficient to train a reliable DGM. In this context, we show that the \( \eta \)-map can be used to perform super-resolution while simultaneously correcting the statistics of low-resolution (LR) samples generated by a high-fidelity DGM trained on sufficient LR data. As a consequence, the corrected super-resolved samples obtained through \( \eta \)-learning and a trustworthy LR DGM exhibit superior statistical characteristics compared to those generated by an HR DGM trained without extreme events. A schematic description of this workflow and a detailed experimental setup are provided in the SI.

\begin{figure}
    \centering
    \includegraphics[width=.99\linewidth]{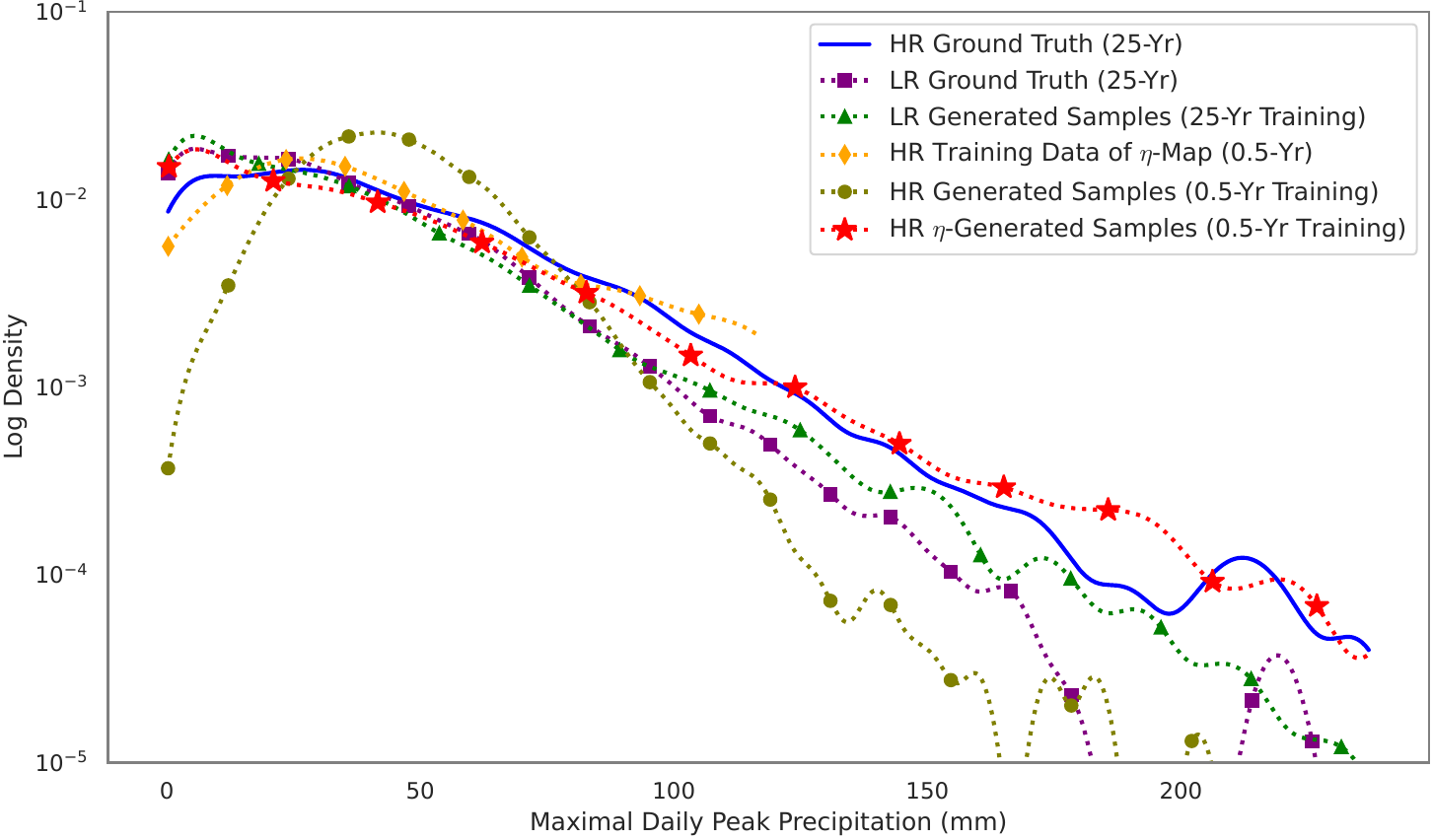}
    \caption{Probability densities of low-resolution (LR) and high-resolution (HR) maximal daily peak precipitation under different processing methods. Solid blue curve: 25-year ground-truth HR data. Dotted purple curve with square markers: 25-year ground-truth LR data. Dotted green curve with triangle markers: 25-year LR samples freshly generated from a Flow Matching (FM) model trained with the 25-year ground-truth LR data. Dotted orange curve with rhombus markers: 0.5-year HR data used to train the HR FM model. Dotted olive curve with circle markers: 9044 HR samples freshly generated from that HR FM model trained with 0.5-year data. Dotted red curve with star markers: HR samples obtained by super-resolving those 25-year LR samples generated from the LR FM model through the $\eta$-map.}
    \label{fig:eta-fm-pdf-compare}
\end{figure}

We make the following observations from Fig. \ref{fig:eta-fm-pdf-compare}. First, it is evident that both the LR FM model trained with 25 years of data and the HR FM model trained with only 0.5 years of data exhibit undesirable performance. The LR model struggles because LR data intrinsically fail to capture the fine-scale details that characterize extreme events, while the HR model performance is primarily hampered by the limited data availability. The latter point is further underscored below. In contrast, the $\eta$-map, despite being trained on only 0.5 years of data, effectively lifts the distribution---particularly its tail---of the LR FM model's samples so that they closely align with the ground truth distribution. The tail elevation relative to the ground truth reflects uncertainty from drawing fresh 25-year LR samples from the LR FM model, rather than direct evidence of over-regularization of the \(\eta\)-map. This result highlights the versatility of the statistical regularization incorporated within the $\eta$-learning framework. Moreover, the visualizations in Supplementary Fig. 10 demonstrate that the $\eta$-map can generate physically plausible extreme events from under-resolved LR samples, a much desired feature for applications in statistical downscaling.

To verify that the shortcomings of \( p_{\text{hi}} \) stem from limited training data, we gradually increase the HR training data to 2.5, 10, and 25 years. An individual FM model is trained for each case using similar protocols, and for each model, 25-year samples are generated freshly using the procedure described in the SI. The resulting observable PDFs are presented in Supplementary Fig. 11. As the training dataset grows, the bulk of the distribution shifts progressively toward the ground truth, and the tail becomes heavier. With the full dataset, the decay rate of the tail eventually aligns with that of the ground truth, as well as with the tail of the push-forward of the $\eta$-map. It is important to note that these models trained with a relatively sufficient amount of HR data are not intended as direct competitors to our method, since the $\eta$-map is trained using only 0.5 years of data. Instead, they serve to clarify the behavior of the true competitor—namely, the FM model trained with 0.5 years of HR data.

\subsection*{$\eta$-Downscaling with Hypothesized Heavier-Tailed Distributions}

\begin{figure}
    \centering
    \includegraphics[width=.99\linewidth]{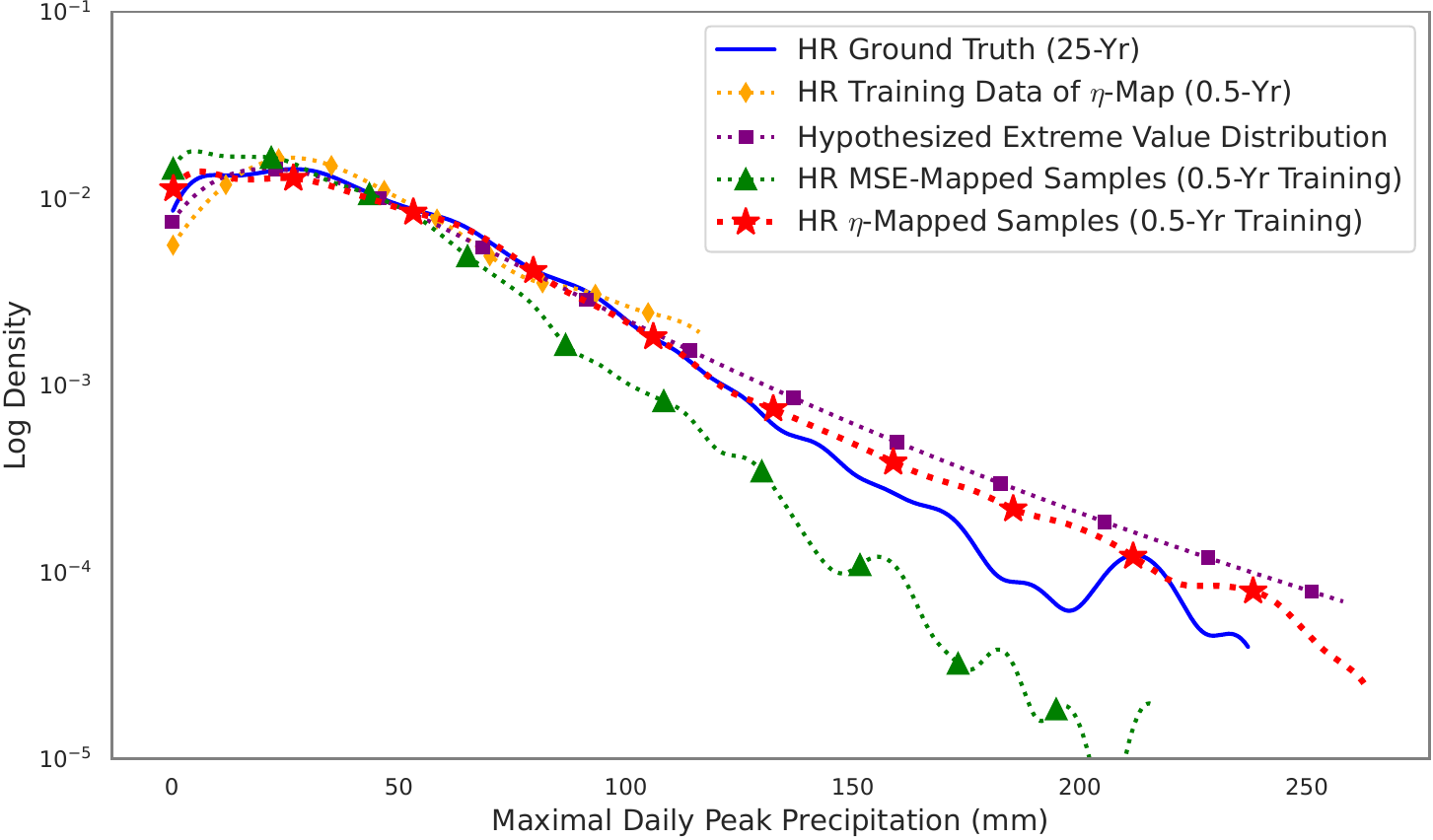}
    \caption{Probability densities of high-resolution (HR) daily maximal precipitation under different processing methods. Solid blue curve: 25-year ground-truth HR data. Dotted orange curve with rhombus markers: 0.5-year HR training data. Dotted purple curve with square markers: hypothesized generalized extreme value distribution (GEVD) used to train the $\eta$-map. Dotted green curve with triangle markers: HR samples obtained by super-resolving 25-year low-resolution (LR) samples through the MSE map. Dotted red curve with star markers: HR samples obtained by super-resolving 25-year LR samples through the $\eta$-map trained with the hypothesized GEVD.}
    \label{fig:pdf-gevd}
\end{figure}

In this last problem, the setup remains identical to the first downscaling example, except that $\nu_0$ is not chosen as $y_\#\mu$, but rather a hypothesized generalized extreme value distribution (GEVD). This minor modification marks a crucial verification of the wide applicability of $\eta$-learning. Specifically, it demonstrates that \( \eta \)-learning is not confined to the ideal case; it can effectively capture potential extreme events that have neither been encountered nor observed and may not be well-represented in a finite dataset. This capability is particularly relevant in addressing challenges posed by the worsening impacts of climate change \cite{seneviratne2021weather}, among other applications. The procedure for obtaining the GEVD is specified in the SI. The resulting GEVD is presented as the dotted purple curve in Fig. \ref{fig:pdf-gevd}. This distribution serves as \( \nu_0 \) used in this analysis. We emphasize that here, $\nu_0$ is a deliberately heavier-tailed reference law used to explore what HR precipitation fields are statistically consistent with the paired LR-HR training data under a specified adverse-tail hypothesis.

The visualizations of the resulting downscaled fields and the corresponding observable PDFs are presented in Supplementary Fig.~12 and Fig.~6. The observable law of the \(\eta\)-downscaled fields is shifted toward the hypothesized GEVD and therefore has a heavier tail than the true HR law. This experiment shows that the framework can generate fields conditional on tail information extending beyond the observed record, thereby broadening our understanding of potential extreme scenarios by leveraging only hypothetical tail information---a critical advantage in risk-sensitive applications.

At the same time, the experiment does not validate the hypothesized $\nu_0$ or the amplified precipitation values produced under that hypothesis. This distinction is important: since the distributional regularizer is designed to enforce the supplied tail statistics, if the tail is misspecified, the learned map can generate correspondingly misspecified extreme amplitudes. We quantify this effect through the controlled sensitivity analysis reported in SI Table~2 as well as Supplementary Fig. 13 and 14, with details provided in the corresponding SI subsection. The supplied tail quantiles are perturbed by a scalar parameter \(\alpha\), for which the misspecification magnitude is proportional to \(|\alpha|\). As the target is moved in the heavier-tail direction from \(\alpha=-1\) to \(\alpha=2\), the full-grid $\ug$-RMSE increases from \(2.926\) to \(3.527\), a \(20.55\%\) increase. $\ug$-$\overline{\mathrm{SSIM}}$ remains between \(0.941\) and \(0.947\), indicating that the main degradation is in amplitude-sensitive pointwise fidelity rather than in the spatial morphology measured by SSIM.

Accordingly, Supplementary Fig.~12 should be interpreted as showing scenarios conditional on the hypothesized GEVD. The fields illustrate how the learned downscaling map responds when an assumed heavier tail is imposed, rather than providing evidence that the amplified extremes will occur or that their magnitudes are physically validated. The experiments therefore demonstrate both the scenario-generation capability of the framework and its dependence on the accuracy of the supplied reference information. Such conditional scenarios can help inform risk assessment and decision-making when the reference tail is treated as an explicit modeling hypothesis and accompanied by proper sensitivity analysis.

\section{Discussion}\label{sec:conclude}
Several theoretical and practical aspects of the framework merit further discussion. First, the constants \(\tilde{C}\) and \(\hat{C}\) in Definition \ref{defn:data-consistent} encapsulate the degree to which a particular estimator fails to generalize effectively to the unseen extreme region \(E\). The exact forms of \(\tilde{C}\) and \(\hat{C}\) are expected to be highly problem-dependent, influenced by factors such as the complexity of \(\Fc\), the smoothness of the true mapping \(y\), the size of the dataset \(n\), and the geometry of the extreme set \(E\). While it is beyond the scope of this work to characterize these constants for general approximation classes, they can be computed explicitly in simple settings. We provide such an example in the SI. They remain a technical component in our theory to convey the following idea: a data-consistent estimator performs better in areas with abundant data while struggling in regions where data is nonexistent. This condition underscores the inherent difficulty of accurately approximating the true mapping in parts of the input space that are underrepresented, especially in the presence of extreme events.

Although determining $\tilde{C}$ and $\hat{C}$ is not central to this work, we make the following comments on their quantitative properties. Based on classical statistical learning theory and the analytic example provided in the SI, we hypothesize that \(\tilde{C}\) and \(\hat{C}\) depend on the aforementioned elements. More relevant to our setting, the signed-error constant \(\tilde{C}\) should be less than \(1\). After all, the defining attribute of our problem is that, in the absence of extreme-event data, the signed approximation error over the non-extreme region \(\Xc\setminus E\) is dominated by the corresponding error over the extreme region \(E\). That is, $|\int_{\Xc\setminus E}(y-\yf)\,\mathrm{d}\mu| <|\int_E(y-\yf)\,\mathrm{d}\mu|$, which is precisely the regime encoded by \(\tilde{C}<1\). By contrast, \(\hat{C}\) functions to quantify the relative size of the absolute error over \(\Xc\setminus E\) compared with that over \(E\); it suffices for \(\hat{C}\) to be finite.

A separate practical concern is the specification of the reference measure $\nu_0$, which is central to the formulation of $\eta$-learning. This raises a natural question: how can $\nu_0$ be determined in practice? While at first glance the approach might appear ad hoc, it is fundamentally grounded in the analytical properties of extensive phenomena. For instance, \(\alpha\)-stable distributions have long been used to model earthquake strength and rainfall amounts \cite{nolan2020univariate}, while max-stable distributions are widely applied to model maximum sea levels and financial risks \cite{coles2001introduction}. Additionally, analytic techniques such as asymptotic analysis for weakly nonlinear systems \cite{Tayfun1980, Weinan1999, Soong_Grigoriou93}, white noise approximations for systems under stochastic excitation \cite{Sobczyk91, Belenky2019}, and the Gibbs hypothesis \cite{Sullivan2009, lecun06, Zhu2019} further exemplify how specific statistical methodologies inform the selection of \(\nu_0\).

When no such theoretical approximations are available, a faithful $\nu_0$ can still be obtained through other means. One increasingly relevant case is when an unlabeled dataset, unsuitable for supervised learning, is available. In such settings, unsupervised learning techniques may be employed to estimate a faithful reference distribution. This perspective highlights an additional advantage of our framework: it naturally bridges supervised and unsupervised learning. Even when none of the arguments above apply, as demonstrated in the last numerical example, one can hypothesize a distribution that describes extreme events, e.g. a generalized extreme value distribution, and understand what type of realizations would be consistent with the underlying dataset and the assumed statistical information. These examples demonstrate how established statistical principles guide the choice of \(\nu_0\), thereby anchoring the \(\eta\)-learning framework in a diverse array of natural and engineered systems. The referenced works provide numerous instances across various scientific and engineering domains, highlighting the broad applicability and versatility of the proposed approach.

Beyond specifying the reference measure \(\nu_0\), one must also decide how to quantify the discrepancy between the learned observable distribution and \(\nu_0\). In this regard, \(W_1\) is attractive not only because of its theoretical optimality in controlling tail errors, but also because it offers practical advantages over alternatives such as KL divergence and the \(L^1\)-log metric \cite{mohamad2018sequential}. The key issue with the usual forward KL divergence, $D_{\mathrm{KL}}(p\|q) = \int p(y)\log\frac{p(y)}{q(y)}\,dy,$ is that discrepancies in the tail are weighted by \(p(y)\), which is small precisely in rare-event regions. As a result, even substantial disagreement between \(p\) and \(q\) in the tail can contribute weakly to the total KL objective. In the present setting, this is undesirable because the main goal is to recover statistically meaningful behavior in low-probability extreme regions.

The \(L^1\)-log metric mitigates this issue by comparing log densities directly. It is therefore more explicitly tail-aware and could in principle serve as a regularizer for $\eta$-learning. However, this advantage comes with a practical limitation: the metric requires reliable density estimates and becomes numerically unstable when either density is zero or extremely small. In particular, when the supports of the learned observable distribution and the reference distribution do not fully overlap, the log-density discrepancy diverges. This issue is especially common in the data-scarce regime considered here, where the learned model may initially assign little or no probability mass to extreme values. In such cases, KL-based objectives and the \(L^1\)-log metric require additional numerical devices to remain well defined.

In contrast, \(W_1\) avoids this support-mismatch obstruction: for probability measures with finite first moments, it gives a finite and stable discrepancy even when the learned and reference observable laws have poorly overlapping tails. A \(W_1\)-based regularizer thus provides a stable training signal in this regime. The practical usefulness of \(W_1\) also comes from the fact that, in one dimension, \(W_1\) admits the quantile representation \eqref{eq:w1-quantile}, which makes the discrepancy straightforward to estimate from samples. More importantly, this representation naturally allows the numerical approximation to be concentrated on the tail; see Methods for details. As a result, $W_1$ allows for a numerically robust mechanism for making the regularization tail-aware.

Despite the theoretical and practical advantages of the $W_1$-based formulation described above, $\eta$-learning remains subject to several limitations that motivate future work. A central limitation is that matching the distribution of a prescribed observable does not identify the dynamical mechanism that generates extreme events. The \(W_1\) regularizer constrains observable statistics, but does not by itself determine the temporal evolution, spatial organization, or broader physical coherence of the generated extremes. To the extent that these properties are recovered, they are inherited from the supervised input-output data rather than supplied by the \(W_1\) term. Consequently, when distinct mechanisms or state configurations induce the same observable law, they are not distinguishable under the present objective. An important direction is therefore to extend the current framework with temporal and spatial constraints, conservation laws, or other physical priors.

This non-identifiability also arises at the level of individual events: the prescribed observable law does not generally determine the precise location, occurrence time, or full spatial structure of an unobserved extreme. Variability in these features across independently trained \(\eta\)-maps should therefore be interpreted as residual uncertainty. Importantly, this uncertainty can be exploited systematically. Ensembles of \(\eta\)-maps can generate diverse yet statistically consistent candidate extremes, which can then be organized according to physically meaningful attributes such as event location, occurrence time, or spatial pattern. Representative scenarios can guide subsequent rounds of adaptive data acquisition, such as targeted high-fidelity simulations or experiments, allowing the corresponding hypotheses to be tested, refined, or rejected while progressively reducing uncertainty in data-scarce regions.

Another limitation is the reliance on an accurate reference law \(\nu_0\). Because the tail-emphasized \(W_1\) estimation is designed to match the supplied tail, misspecification of \(\nu_0\), or a shift in the target tail relative to \(\nu_0\), can be transferred directly to the generated extremes. A promising extension is a distributionally robust formulation that replaces a single reference law with an ambiguity set of plausible laws, with particular emphasis on uncertainty and shifts in the tail. Worst-case or uncertainty-weighted tail objectives could reduce overcommitment to any single hypothesized distribution while preserving awareness of extreme behavior.

Finally, the present theory and algorithms are developed for scalar observables, for which \(W_1\) enables efficient computation and underpins the tail optimality results. For a vector observable \(G=(g_1,\ldots,g_r)\), one may either introduce a physically meaningful severity function \(s:\mathbb R^r\to\mathbb R\) and apply the present framework to \(g=s\circ G\), or use a sum of componentwise \(W_1\) penalties when matching the marginal laws is sufficient. The latter scales linearly with \(r\), but does not constrain the correlation structure among the components. When the joint law of \(G\) is essential, genuinely multivariate discrepancies based on, for example, sliced Wasserstein distance or entropically regularized optimal transport are required \cite{bonneel2015sliced, cuturi2013sinkhorn}. Extending the present tail-focused theory and scalable algorithms to these settings is another important direction.

Taken together, the theoretical and numerical results position $\eta$-learning as a general mechanism for incorporating partial knowledge of extreme-event statistics into data-driven models when representative extreme samples are unavailable. This capability is particularly relevant in risk-sensitive settings where conventional training data provide poor coverage of consequential rare events. Beyond the examples considered here, the framework may support broad applications in agent-based decision-making, synthetic financial data generation, safety-critical control, climate-extreme modeling, and other domains in which reliable characterization of extreme events is essential.

\section{Methods}
The optimization formulation of $\eta$-learning in \eqref{eq:geneces} and \eqref{eq:geneces-u} serves as the foundation of our method. However, several numerical considerations must be addressed when applying it in practice. We focus solely on the case of \eqref{eq:geneces-u}, as \eqref{eq:geneces} can be viewed as a special case by setting $u$ as the identity map and $y=g$ as the unknown.

\subsection*{Estimating $W_1$ to Emphasize Tails}
A central computational component of $\eta$-learning is the distributional regularizer $W_1(\cdot,\nu_0)$. While Wasserstein distances are generally expensive to compute in high dimensions, our setting only requires $W_1$ in one dimension (over the observable), which admits an equivalent quantile representation. Concretely, for $\nu_1,\nu_2 \in \Pc_1(\Yc)$ with quantile functions $F^{-1}_{\nu_1}$ and $F^{-1}_{\nu_2}$, we have
\begin{equation}
    \label{eq:w1-quantile}
W_1(\nu_1,\nu_2) = \int_0^1 \left| F_{\nu_1}^{-1}(q)-F_{\nu_2}^{-1}(q) \right|\,dq,
\end{equation}
which can be easily shown by applying the Fubini's Theorem to Theorem 1.7 in \cite{sot}. Applying this formula to $W_1((g \circ \phi)_\#\mu, \nu_0)$ yields
\[
W_1((g \circ \phi)_\#\mu, \nu_0) = \int_0^1 \left|F_{(g \circ \phi)_\#\mu}^{-1}(q) - F_{\nu_0}^{-1}(q)\right| \, \mathrm{d}q.
\]
We approximate this integral by Monte Carlo quadrature over a set of quantile levels $\Qc=\{q_i\}_{i=1}^{n_q}$:
\begin{equation}
W_1((g\circ \phi)_\#\mu,\nu_0)\approx \frac{1}{n_q}\sum_{i=1}^{n_q}\left|F_{(g\circ \phi)_\#\mu}^{-1}(q_i)-F_{\nu_0}^{-1}(q_i)\right|.
    \label{eq:w1-monte-carlo}
\end{equation}
The reference quantiles $F_{\nu_0}^{-1}(q_i)$ are assumed tractable. The model-dependent quantiles $F_{(g\circ \phi)_\#\mu}^{-1}(q_i)$ are estimated from an auxiliary dataset $\Dc_{\tilde{\bfx}}=\{\tilde{\bfx}_j\}_{j=1}^{n_{\tilde{\bfx}}}$, with $\tilde{\bfx}_j\sim\mu$, by evaluating $g(\phi(\tilde{\bfx}_j))$ and taking empirical order statistics. Note that $n_{\tilde{\bfx}}$ is in general much larger than $n$, the size of the training dataset, for accurate estimation.

While the Monte Carlo estimation in \eqref{eq:w1-monte-carlo} is theoretically sound, it presents a significant limitation in our context. Since we require a highly accurate approximation to the higher quantiles after training, if the probability values $\Qc$ were sampled uniformly, higher quantiles will be significantly underrepresented during the optimization. This results in insufficient emphasis on the tail. To prioritize extreme behavior, we sample $\Qc$ from a proposal distribution skewed towards 1 (i.e. a tilted measure), so that the $W_1$ term concentrates its signal on the tail. A simple yet effective choice of the tilting is to define a probability cutoff \(\tau\) and estimate
\[
\frac{1}{1-\tau} \int_\tau^1 \left|F_{(g\circ\phi)_\#\mu}^{-1}(q) - F_{\nu_0}^{-1}(q)\right| \, \mathrm{d}q
\]
as a proxy for $W_1((g\circ \phi)_\#\mu, \nu_0)$. In practice, $\tau$ is typically chosen around 0.95 or 0.975.

This approach may raise concerns about neglecting the bulk of the distribution. However, we argue that this issue is mitigated because non-extreme regions are well-represented in the training data by assumption. These non-extreme events are thus adequately addressed through the ERM loss. Consequently, it is the tail should receive the primary attention through the $W_1$ term to align with the central goal of the regularization. To sum up, the practical optimization objective of \eqref{eq:geneces-u} is
\begin{equation}
    \min_{\phi\in\Fc} \frac{1}{n} \sum_{i=1}^n\ell(\phi, \bfx_i, \bfu_i) + \frac{\lambda}{n_q} \sum_{i=1}^{n_q} \left|F_{(g\circ\phi)_\#\mu}^{-1}(q_i) - F_{\nu_0}^{-1}(q_i)\right|,
    \label{eq:eta-u-practice}
\end{equation}
where $\{q_i\}_{i=1}^{n_q}$ are drawn from $[\tau, \,1]$.

\subsection*{Effective Training Strategies}
Several difficulties arise in the actual optimization of \eqref{eq:eta-u-practice}. We introduce the following shorthand notation for subsequent discussions:
\begin{equation*}
    \Lc_\theta^{(k)}(\phi_\theta^{(k)};\Dc, \Qc,\nu_0, g) := \frac{1}{n} \sum_{i=1}^n\ell(\phi_\theta^{(k)}, \bfx_i, \bfu_i) + \frac{\lambda}{n_q} \sum_{i=1}^{n_q} \left|F_{\left(g\circ{\phi_\theta^{(k)}}\right)_\#\mu}^{-1}(q_i) - F_{\nu_0}^{-1}(q_i)\right|,
\end{equation*}
where the superscript $(k)$ denotes the iteration index in the optimization process. Note that the ERM component of $\Lc^{(k)}_\theta(\phi_\theta^{(k)}; \Dc, \Qc, \nu_0, g)$ can be computed in a mini-batch fashion, consistent with standard machine learning practices, and the tail loss component can be similarly handled (although often not necessary since $|\Qc|$ is on the order of 100 in the experiments we conduct). We also assume $\Fc = \Fc_\theta$ is a neural network optimized with a stochastic first-order optimization method $\Mc$ (such as Adam \cite{kingma2014adam}). The update rule is denoted by
\[
\phi_\theta^{(k+1)}
\leftarrow
\Mc\left(\phi_\theta^{(k)}, \nabla_\theta \Lc_\theta^{(k)}\right).
\]
The gradients \(\nabla_\theta \Lc_\theta^{(k)}\) can be efficiently computed using automatic differentiation libraries available in software such as PyTorch \cite{paszke2019pytorch}.

\paragraph*{\textit{Memory Challenges and Tail-Set Selection for \(W_1\) Estimation.}}
Even with the tail-focusing proxy of $W_1$ discussed above, accurate estimation of the higher quantiles still requires evaluating \(g\circ\phi_\theta\) over the auxiliary set $\Dc_{\tilde{\bfx}}$, which is typically much larger than the labeled training set. If this full auxiliary evaluation were included in every gradient-carrying update, the memory cost would be substantial.

To reduce this cost, we exploit the fact that the empirical quantile loss \eqref{eq:w1-monte-carlo} requires only the \(n_q\) inputs from $\Dc_{\tilde{\bfx}}$ whose current observable values realize the quantile levels in $\Qc$. We therefore develop an inference-informed tail-set selection procedure. Periodically, we run inference on the full auxiliary set $\Dc_{\tilde{\bfx}}$ and compute
\[
\left\{g(\phi_\theta(\tilde{\bfx}_j))\right\}_{j=1}^{n_{\tilde{\bfx}}}.
\]
We then identify the indices of the inputs corresponding to the prescribed quantile levels in $\Qc$ and store them in a set $\Ic$. Between refreshes, the \(W_1\) loss is evaluated only on the $\Ic$-indexed inputs. Thus, full inference on $\Dc_{\tilde{\bfx}}$ is used only to identify the relevant tail samples, while the gradient-carrying \(W_1\) update is restricted to the selected quantile samples. The refresh frequency is controlled by a hyperparameter $\omega$, with $\Ic$ updated every $\omega$ iterations. This procedure substantially reduces memory usage while preserving an accurate estimate of the tail contribution to the \(W_1\) loss. A detailed complexity analysis is provided below to demonstrate the benefit.

\paragraph*{\textit{Gradient Conflict and ERM Pre-training.}}
The tail-set selection procedure also explains a potential source of optimization instability. The \(W_1\) loss is evaluated on inputs selected according to the current model-induced observable distribution \((g\circ\phi_\theta)_\#\mu\). Early in training, before the supervised input-output relation has been adequately learned, the ordering of the auxiliary observable values \(g(\phi_\theta(\tilde{\bfx}_j))\) can be unreliable. Consequently, inputs that should correspond to the bulk under an ERM-consistent predictor may be spuriously selected into $\Ic$ and treated as high-quantile samples. The \(W_1\) term then pulls these samples toward upper quantiles of the reference distribution \(\nu_0\), while the ERM loss pulls the model toward the labeled input-output relation. These two signals can be incompatible, leading to oscillatory or unstable optimization.

Our main stabilization mechanism is ERM pre-training. We first minimize the ERM loss alone and initialize \(\phi_\theta\) with the resulting ERM estimator \(\uf\) when solving \eqref{eq:eta-u-practice}. The initial index set $\Ic$ used by the \(W_1\) penalty is also computed from this ERM-pretrained model. Thus, the first \(W_1\) correction is applied to quantile samples selected under a predictor that already captures the observed input-output relation. This substantially reduces the chance that bulk samples are spuriously treated as tail samples because of unreliable early predictions. During \(\eta\)-learning, $\Ic$ is periodically refreshed as \(\phi_\theta\) evolves, keeping the \(W_1\) correction aligned with the current observable distribution.

\paragraph*{\textit{Choice of \(\lambda\).}}
The balancing parameter \(\lambda\) controls the relative magnitude of the ERM and \(W_1\) gradient components. Writing
\[
\Lc_\theta
=
L_{\mathrm{ERM}}(\theta)
+
\lambda L_{W_1}(\theta),
\]
we have
\[
\nabla_\theta \Lc_\theta
=
\nabla_\theta L_{\mathrm{ERM}}(\theta)
+
\lambda \nabla_\theta L_{W_1}(\theta).
\]
A practical scale for \(\lambda\) is obtained by choosing it so that the scaled \(W_1\) gradient has a comparable Euclidean norm to the ERM gradient during the early \(\eta\)-learning iterations:
\[
\lambda \|\nabla_\theta L_{W_1}(\theta)\|_2 =
\|\nabla_\theta L_{\mathrm{ERM}}(\theta)\|_2 \implies
\lambda
=
\frac{
\|\nabla_\theta L_{\mathrm{ERM}}(\theta)\|_2
}{
\|\nabla_\theta L_{W_1}(\theta)\|_2+\varepsilon
},
\]
with a small \(\varepsilon>0\) for numerical stability. This gradient-norm balancing rule can also be applied adaptively during training, helping prevent the \(W_1\) correction from overwhelming the supervised anchor.

However, \(\lambda\) only rescales the \(W_1\) gradient; it does not change its direction. Therefore, while magnitude balancing can be beneficial, it alone cannot resolve the directional incompatibility caused by unreliable tail-set selection. For this reason, ERM pre-training and periodic tail-set refresh are the primary mechanisms for stable optimization, while \(\lambda\) serves as a secondary balancing parameter that controls the strength of the residual tail correction. This interpretation is consistent with the robustness experiment in Supplementary Fig.~4, where the learned observable distribution remains stable as \(\lambda\) varies from \(10^{-4}\) to \(10\) while all other training configurations are fixed.

\paragraph*{\textit{Structural Challenges from the Observable.}} The structure of the observable \(g\) can also complicate training, particularly when \(g\) depends on only a small subset of the components of \(\bfu \in \mathbb{R}^m\), with the relevant components varying across data points. A canonical example is the maximum function \(\max(\cdot)\), a cornerstone of extreme value theory \cite{coles2001introduction}, for which \(g(\bfu)\) depends solely on the maximal entry of \(\bfu\). The index of this entry may change as \(\phi_\theta\) is updated, inducing oscillations and hindering convergence of \((g \circ \phi)_{\#}\mu\). To mitigate this issue, we extend the tail-selection procedure by tracking the indices of the components of \(\bfu\) that determine the output of \(g\). These indices are collected in a set \(\Jc\), which is updated alongside \(\Ic\) every \(\omega\) iterations.

Together, these considerations lead to the Inference-Informed Continual Training (IICT) procedure summarized in Algorithm \ref{alg:train}, with the index-set update detailed in Algorithm \ref{alg:update}. In summary, IICT initializes the model with the ERM estimator $\uf$ and periodically refreshes two index collections: $\Ic$, which identifies the auxiliary inputs used to evaluate the quantile loss efficiently, and, when applicable, $\Jc$, which tracks the active state components determining the observable. Thus, $\Ic$ provides the general memory reduction used throughout the framework, whereas $\Jc$ supplies additional, problem-specific stabilization when required by the structure of $g$.

\begin{algorithm}[!t]
\caption{$\eta$-Learning: Inference-Informed Continual Training (IICT)}\label{alg:train}
\begin{algorithmic}[1]
\Require $\uf, \Dc, \Dc_{\tilde{\bfx}}, \Qc, \nu_0, \omega, \lambda, \Mc, N$
\State $k \leftarrow 0$
\State{$\phi_\theta^{(k)} \leftarrow \uf$} \Comment{Initialize $\phi_\theta^{(k)}$ with ERM estimator}
\State{$\Ic, \Jc \leftarrow$ \texttt{Update\_Tail\_Index}($\phi_\theta^{(k)}, \Qc, \Dc_{\tilde{\bfx}}$)} \Comment{Initialize tail sample index sets}
\State
\While{$k < N$} \Comment{Start training}
        \State{\(\nabla_\theta \Lc_\theta^{(k)} = \nabla_\theta \Lc_\theta(\phi_\theta^{(k)}, \Ic, \Jc;\Dc, \Qc,\nu_0, g)\)}
        \State{$\phi_\theta^{(k+1)} \leftarrow \Mc(\phi_\theta^{(k)},\nabla_\theta \Lc_\theta^{(k)})$} \Comment{Update $\phi_\theta^{(k)}$ with $\eta$-loss}
        \State $k \leftarrow k+1$
        \If{$k \mod \omega = 0$}
            \State $\Ic, \Jc \leftarrow$ \texttt{Update\_Tail\_Index}($\phi_\theta^{(k)}, \Qc, \Dc_{\tilde{\bfx}}$)
        \EndIf
\EndWhile
\State \Return $\ug = \phi_\theta^{(N)}$
\end{algorithmic}
\end{algorithm}

\begin{algorithm}[H]
\caption{Update Tail Index}\label{alg:update}
\begin{algorithmic}[1]
\Require $\phi_\theta, \Qc, \Dc_{\tilde{\bfx}}$

\State $\Dc_{\tilde y}, \Ic, \Jc \leftarrow \emptyset$ \Comment{Initialization}
\For{$i=1:n_{\tilde{\bfx}}$} \Comment{Run inference on $\Dc_{\tilde{\bfx}}$}
        \State{$\Dc_{\tilde y} \leftarrow \Dc_{\tilde y} \cup \{g(\phi_\theta(\tilde{\bfx}_i))\}$}
\EndFor
\State{$\pi \leftarrow$ \texttt{Sorted}($\Dc_{\tilde y}$)} \Comment{Ascending sort; return permutation}
\For{$q \in \Qc$} \Comment{Start updating indices}
        \State{$y_\theta^{(q)} \leftarrow F_{\left(g \circ \phi_\theta\right)_{\#} \mu}^{-1}(q)$} \Comment{Evaluate observable quantile}
        \State{$i_q \leftarrow \pi^{-1}(\lceil qn_{\tilde{\bfx}}\rceil)$} \Comment{Identify the input index for the $q$-th quantile}
        \State{Find $j_q$ s.t. $g\left(\left[\phi_\theta\left(\tilde{\bfx}_{i_q}\right)\right]_{j_q}\right) = y_\theta^{(q)}$} \Comment{Determine the dimensions that trigger $g$ \\ \hfill if applicable}
        \State $\Ic \leftarrow \Ic \cup \{i_q\}$  \Comment{Update $\bfx$-indices}
        \State{$\Jc \leftarrow \Jc \cup \{j_q\}$} \Comment{Update $\bfu$-indices}
\EndFor
\State \Return $\Ic, \Jc$
\end{algorithmic}
\end{algorithm}

\subsection*{Computational Complexity}

Here, a detailed complexity analysis of the full $\eta$-Learning algorithm, i.e. Algorithm 1, is provided. We distinguish the inference-only computation used to identify the relevant tail samples from the gradient-carrying computation used to update the neural network. Let \(C_{\mathrm{f}}\) and \(C_{\mathrm{b}}\) be the per-sample forward- and backward-propagation costs through \(g\circ\phi_\theta\), respectively. Repeated input indices are allowed when different quantile levels correspond to the same empirical sample; hence, the gradient-carrying $W_1$ loss always contains \(n_q\) samples.

We consider the conservative case \(\omega=1\), in which case the quantile indices are updated at every training iteration. First, \(g\circ\phi_\theta\) is evaluated on all \(n_{\tilde{\bfx}}\) inputs in inference mode, and the resulting scalar values are sorted to identify the prescribed quantiles. This step has cost
\[
O\left(
n_{\tilde{\bfx}}C_{\mathrm{f}}
+
n_{\tilde{\bfx}}\log n_{\tilde{\bfx}}
\right).
\]
Because this full-sample evaluation is forward-only, it does not require retaining a computational graph. The \(W_1\) loss is then computed and back-propagated using only on the \(n_q\) quantile-indexed samples, with cost
\[
O\left(
n_q(C_{\mathrm{f}}+C_{\mathrm{b}})
\right).
\]
Therefore, the additional per-iteration cost of the \(W_1\) regularization is
\[
O\left(
n_{\tilde{\bfx}}C_{\mathrm{f}}
+
n_{\tilde{\bfx}}\log n_{\tilde{\bfx}}
+
n_q(C_{\mathrm{f}}+C_{\mathrm{b}})
\right).
\]
The main practical distinction is that forward propagation is performed on the full input pool, whereas the more expensive backward propagation is restricted to the \(n_q\) dynamically selected tail samples. Note that $n_{\tilde{\bfx}}$ and $n_q$ can be mini-batched if the memory cap is of concern.

In the precipitation experiment, \(n_{\tilde{\bfx}}=9044\), and the matched tail begins at approximately the \(0.975\) quantile. The quantile grid therefore contains approximately
\[
n_q \simeq (1-0.975)n_{\tilde{\bfx}}\simeq 226
\]
terms. Thus, although all \(9044\) inputs are used in the inference-only forward pass, only approximately \(2.5\%\) as many samples are used in the gradient-carrying \(W_1\) update.

For the Flow Matching (FM) experiment, the LR FM model is trained on the LR data and used to generate the 25-year-equivalent LR samples exactly as it would be without \(\eta\)-learning. The trained \(\eta\)-map is then applied once to each generated LR sample to obtain a tail-corrected HR sample. For fair comparison, $n_{\tilde{\bfx}}$ LR samples are generated, and the additional \(\eta\)-evaluation cost is therefore
\[
O\left(n_{\tilde{\bfx}}C_{\mathrm{f}}\right).
\]
This is a single inference-only post-processing pass per sample. It does not alter the training cost of the FM model, does not add computations to its numerical sampling steps, and requires no \(W_1\) evaluation or backward propagation. The other additional expense is the separate, one-time training of the \(\eta\)-map, whose per-iteration complexity has been given above.

\subsection*{Experimental Details}
Complete experiment-specific details—including the analytic maps for the toy examples, precipitation data preprocessing, training--testing splits, construction of $\nu_0$ and $\Qc$, tail cutoffs, auxiliary sample sizes, network architectures, optimization hyperparameters, values of $\lambda$ and $\omega$, training durations, baselines, and evaluation procedures—are provided in the Experimental Setups and Details section of the SI.

\backmatter

\subsection*{Data availability}
The data used in this study are available from Zenodo under the Creative Commons Attribution 4.0 International (CC BY 4.0) license at \url{https://doi.org/10.5281/zenodo.21635446}. The trained models and generated samples are available separately from Zenodo under the CC BY 4.0 license at \url{https://doi.org/10.5281/zenodo.21635468}.

\subsection*{Code Availability}
The source code used to produce the results reported in this study is publicly available under the MIT license at \url{https://github.com/kai-ovo/eta} and is permanently archived in Zenodo at \url{https://doi.org/10.5281/zenodo.21636362}.

\backmatter


\section*{Acknowledgements}
We thank Jennifer Zheng (Stanford ICME) for many helpful discussions and feedback on the theoretical developments.

\section*{Funding Statement}
This work was supported by the Vannevar Bush Faculty Fellowship (grant no. VBFF N000142512059) and the AFOSR (grant no. FA9550-23-1-0517).

\section*{Author Contributions Statement}
K.C. and T.P.S. conceived the research and developed the methodology. K.C. implemented the methods, conducted the numerical experiments, analyzed the results, prepared the figures, and wrote the original draft. T.P.S. supervised the research, acquired funding, analyzed the results, and reviewed and edited the manuscript. Both authors discussed the results and approved the final manuscript.

\section*{Competing Interests Statement}
The authors declare no competing interests.

\clearpage
\includepdf[pages=-,pagecommand={\thispagestyle{empty}}]{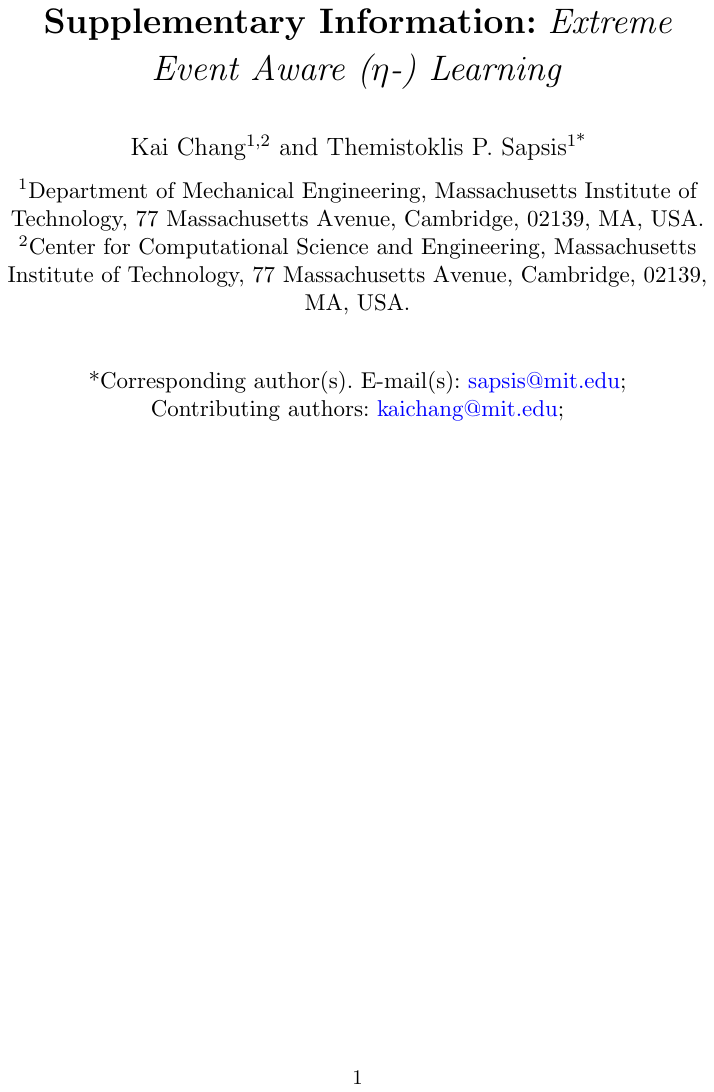}

\end{document}